\documentclass[11pt,a4paper]{article}

\usepackage[T1]{fontenc}
\usepackage[utf8]{inputenc}
\usepackage[margin=2.5cm]{geometry}
\usepackage{microtype}
\usepackage{amsmath}
\usepackage{libertinus}
\usepackage[libertinus]{newtxmath}
\usepackage[scaled=0.83]{beramono}
\usepackage{graphicx}
\usepackage{booktabs}
\usepackage{tabularx}
\usepackage{multirow}
\usepackage{makecell}
\usepackage{longtable}
\usepackage{threeparttable}
\usepackage[table]{xcolor}
\usepackage{etoolbox}
\usepackage{enumitem}
\usepackage{caption}
\usepackage{placeins}
\usepackage{tikz}
\usetikzlibrary{positioning,arrows.meta,decorations.pathreplacing,calc}
\usepackage[numbers,sort&compress]{natbib}
\usepackage{hyperref}
\usepackage[capitalise,noabbrev]{cleveref}

\setlist{itemsep=2pt,topsep=4pt}
\definecolor{linkblue}{HTML}{2a78d6}
\definecolor{ink}{HTML}{2b2a26}
\definecolor{rowshade}{HTML}{f0efe9}
\hypersetup{colorlinks=true,linkcolor=ink,citecolor=linkblue,urlcolor=linkblue}
\newcommand{\doi}[1]{doi: \href{https://doi.org/#1}{\nolinkurl{#1}}}
\AtBeginEnvironment{table}{\rowcolors{2}{}{rowshade}}
\newcommand{\TC}{\code{TC}}
\newcommand{\vcc}{\code{pi-vcc}}
\newcommand{\pidef}{\code{Pi default}}
\newcommand{\TCng}{\code{TC-noguidance}}
\newcommand{\TCnr}{\code{TC-noreason}}
\newcommand{\code}[1]{\texttt{#1}}
\newcommand{\kept}{\emph{kept}}
\newcommand{\notkept}{\emph{not kept}}
\newcommand{\entry}[1]{\{#1\}}
\newcommand{\repro}{\url{https://github.com/felixwoolford/pcc-replication-package}}

\title{Wait...what was I doing? A Case Analysis of How Compaction Methods Shape a Coding Agent's Continuity in Task and Metacognitive Stance}

\date{}
\author{Felix Woolford, Keisuke Suzuki\\[2pt]
\small Center for Human Nature, Artificial Intelligence, and Neuroscience (CHAIN), Hokkaido University}

\begin{document}
\maketitle

\begin{abstract}

When a coding agent's context window fills, compaction replaces its history with something shorter. Compaction methods have usually been judged by whether relevant information survives or how well a compacted agent performs on benchmark tasks. We instead ask about the goal-oriented stance of the agent: did the agent carry on in its task \textit{as it was doing before} compaction? In the Pi coding agent harness, we compare Pi's default model summary method with two alternative methods: \code{transparent-compact-pi} (\TC), our own design, and \vcc{}, a popular third-party extension which takes a similar approach. We tested them on a natural case discovered in an agentic-coding session. In that session, an agent using a local model (Qwen3.8-27B) identified and settled on a solution to its task. But after its first compaction, before building anything, it changed course and produced a workaround instead. In the primary experiment, we forked that session from the moment just before the compaction and varied the compactor over 35 runs per arm. \TC{} continued the pre-compaction direction significantly more often than both alternatives. With compaction switched off the agent continued it in all 15 runs, supporting the test as a measure of continuity. In two subsequent experiments we asked what drives \TC's advantage. Without its agent-facing guidance on the compacted state, the agent kept its direction somewhat less often, but not significantly so. Without access to the agent's earlier reasoning, which neither alternative method can reach, it still kept its direction significantly more often than under the alternatives. We interpret the results as pointing to the presence of two carriers of an earlier decision across compaction: its explicit statement, in this case stated only in reasoning, and its implicit consequences, the agent's prose and actions that followed the statement. \TC's advantage appears to stem from the fact that it carries \textit{both} more faithfully. This interpretation is further reinforced by a study on a second model (GPT-6.1 Terra), which shows the same greater continuity for \TC. We conclude by framing compaction as a metacognitive problem: it disrupts the agent's knowledge of its own intentions, and what the compacted context keeps of the agent's own earlier words and actions shapes whether it can still recognise its earlier plan.

\end{abstract}

\section{Introduction}
\label{sec:intro}

Interact with a coding agent for long enough and you will inevitably encounter the effects of compaction. When the context window fills, the harness replaces older history with something shorter, usually a summary the model writes, and carries on \citep{rajasekaran2025context,anthropic2026contextwindow}. Compaction is extremely useful, because it allows a working session to run far longer than a single context window can hold; and even when a session does fit within a single context window, it can help in reducing long contexts which can degrade performance \citep{liu2024lost,hong2025contextrot,breunig2025contextsfail}. But following compaction, the agent often appears to change: \textit{``The new Claude is not the old Claude''} \citep{ethayarajh2026ghosts}. User descriptions of issues that follow compaction show a number of recurring trends. Beyond a general drop in quality \citep{paterson2026lobotomizing}, users report agents that drift from plans laid before, revive approaches already rejected, redo finished work, stop following standing instructions, act on plans still awaiting approval, mistake who said what, and lose the tone they had established \citep{gh29890,batool2026compaction,gh18477,gh4017,gh10006,gh10528,gh14941,gh11582,lee2026personality}. 

In the most common form, a compacted summary keeps what a model ``guesses is important'' \citep{anthropic2026contextwindow}. But not all of these user reports describe an important lost fact per se. The agent can still have the plan, the instructions and the files it has written. But what often seems to catch the user's attention is that it changes its \emph{relationship} to those things: 
The history that makes sense of a plan's direction, an inflation or deflation of the relevance of certain facts, and the particular interactive dynamic that the user and agent operate in. It can also change how the agent holds what it knows: whether a question is settled or still open, and how confident it is in each of its earlier claims, sometimes falsely so (\cref{sec:exploratory}). 
To date, evaluations of compaction have mainly asked what information survives, or whether the task can still be completed afterwards (\cref{sec:related}). A retention probe, though, cannot see most of what users report.

\paragraph{}
An agent partway through a task is engaged in a goal-directed activity. In this framing, rather than being simply the history of the conversation with the user, the context is the medium in which the agent's state is held, along with the state of the files that it is operating on. Compaction disrupts that state, and disrupts that goal-directed activity, and to evaluate its recovery from that disruption we should ask whether the activity resumes as it would have gone on had the disruption not occurred at all. Ideally, compaction would change nothing about the agent's relationship to its task. But inevitably information must leave the context, and the question is how does that loss affect the relationship.  

Compaction methods differ first in how they select what leaves the context \citep{wang2026compression}. A model-written summary, such as Pi's default, selects by content; a mechanical compactor selects by structure and, paired with retrieval, leaves selection by content until the agent asks for something (\cref{sec:compacting}). \code{transparent-compact-pi} \citep[\TC;][]{woolford2026transparent,tcpi} and \vcc{} \citep{sting8kvcc} are both mechanical compactors with retrieval in this sense. They differ in the form of the record they leave, in what they tell the agent about its compacted state, and in what their recovery can reach (\cref{sec:systems}).

After compaction the agent also has to come to know its own state again: what it had intended, which of the claims it carries it had established, and what it no longer has. Compaction is therefore also a problem of metacognition, of the agent's knowledge of its own memory, and of inferring its own earlier intentions from what remains (\cref{sec:related,sec:cogsci}).

\paragraph{}
This paper asks a behavioural question, about one aspect of the continuation of the pre-compacted state: \textbf{In one natural case of agentic coding, how do different compaction methods affect whether a coding agent carries on in the direction it was already heading with respect to its task, and what carries that direction across compaction?}

We use a lightweight and open-weight model, Qwen3.8-27B \citep{qwen2026qwen38} run locally, for two reasons:
\textbf{(1)} Lightweight local models are useful for privacy, cost, and customisation, but they are weaker reasoners, and so less robust to what compaction does \citep{woolford2026transparent}: a weaker model's summary leaves more gaps and ambiguities, and the weaker model is less able to reason around them, or, after mechanical compaction, to notice what is missing and search for it. 
\textbf{(2)} Secondly, with this model we can read the reasoning thread in full. Hosted closed-weight models, such as Claude, hide their reasoning or show only a summary of it. With Qwen we can read directly how the agent reflects on its own compacted state: what it believes it had done, what it thinks it lost, and how it decides what to do next.

With that model, we examine a single case, which we found in the history of a working coding session rather than constructing directly for the test. The session was one of many produced by a factory framework, in which one agent develops a project for a ``user'' played by another agent; nothing in it was arranged to make compaction fail (\cref{sec:testbed}). In one such session, working in the Pi coding agent \citep{zechner2026pi}, the model discovered and settled on a way to solve its task exactly as the user had requested. Its first compaction came before it had built anything. Afterwards it went another way, shipped a workaround, and told the user -- wrongly -- that the approach it had previously settled on was impossible. A decision made before compaction was lost in it, and the agent's behaviour changed as a result. We isolated that moment and turned it into a test.

We froze the session at that point and ran it forward many times, changing only the compactor (Experiment~1), and with no compaction at all (Experiment~2), which gives the reference: an agent whose relationship to its task nothing has disturbed. With the \code{transparent-compact-pi} (\TC) compaction method the agent kept its direction about twice as often as under either alternative, and came closest to that reference. 
We then took \TC{} apart in two further experiments. Experiment~3 asked whether \TC's agent-facing guidance text drives that advantage: without it, the agent kept its direction somewhat less often, but not significantly so. Experiment~4 asked whether the advantage depends on \TC's access to the agent's earlier reasoning, which neither alternative can reach: without that access, the agent still kept its direction significantly more often than under the alternatives. Finally, we tested whether this implicit carriage is also present on a different model, GPT-6.1 Terra (Experiment~5), and found that it was.

We argue that the evidence suggests that the decision crosses compaction in two ways: as an explicit statement, here written in the agent's reasoning, and through its implicit consequences, the visible text and actions that followed from it. \TC's advantage in this case appears to be a consequence of the way that it carries both. From the agent's side, the two carriers are two routes by which it comes to ``know'' its own earlier intention (\cref{sec:cogsci}).
An exploratory reading of the agent's post-compaction reasoning shows that keeping the pre-compaction direction tracks this metacognitive element: agents kept their direction mostly when they came back to the relevant code having already recognised their own earlier intention, but often lost it when they had to re-derive it again from the code. 

\section{Background}

\subsection{Compacting information}
\label{sec:compacting}

\paragraph{} \citet{wang2026compression} describe context compression for agents as a pipeline. A selection stage decides which parts of the context are compressed and which are kept exact; a compression mechanism transforms the selected parts; the result is stored, in the working context or in an external store; and a recovery stage later retrieves stored content when it is needed. We use \emph{compaction}, the term used by coding-agent harnesses, for this whole pipeline. Its selection stage takes place in advance, before the rest of the session is known, and the methods compared here differ first in how they make it:

\begin{itemize}
	\item A \textbf{summary} selects \emph{by content} in advance. The model judges which information in this particular session will matter next, and summarisation rewrites it into a shorter text. A summary is therefore a prediction about the session's future needs. It is also a new text: it restates the agent's state in another voice, which can shift what reads as settled, what reads as important, and how cautious the agent should be.

	\item A \textbf{mechanical} compactor selects \emph{by structure} in advance. Without a model call, it keeps, shortens or drops entries by type and position (user messages, the agent's prose, tool calls, the first request, the most recent stretch), under a fixed rule that is the same for every session and so predicts nothing about what comes next. A mechanical compactor can be paired with externalisation and retrieval: the full history is kept in an external store, and selection by content happens later, through a recovery tool, once the agent knows what it needs. Until the agent asks, what it sees is the structurally reduced record, so the form of that record matters too.
\end{itemize}

\paragraph{} A second property is independent of the kind of selection: what the post-compaction context tells the agent about its own compacted state. It can state which content was removed. It can state the status of what remains, for instance that carried-over statements are earlier claims and not verified facts. And it can say where the removed content can be recovered, and how. Harnesses already provide some of this. Pi opens every post-compaction context with ``The conversation history before this point was compacted into the following summary'' \citep{zechner2026pi}. Claude Code's goes further and says \citep{claudecode2026} that the most recent messages are ``kept verbatim'', that the summary ``was written without seeing them'' and may be out of date, and where the full transcript can be read. We refer to this property as \emph{compaction-state guidance}, and compactors differ in how much of it they give. Guidance is not one of the dimensions of \citeauthor{wang2026compression}'s taxonomy, but it bears directly on a risk that they identify for externalisation and retrieval: stored content may be available but not re-accessed, or is re-accessed at the wrong time or incorrectly. An agent that does not know what was removed has little reason to look for it. Guidance gives the agent information about its own memory rather than about the task, and it is the operational form of a metacognitive argument we develop in \cref{sec:cogsci}.

\subsection{Related work}
\label{sec:related}

\paragraph{Evaluating compaction's effects.} In practice with coding agents, compaction is mainly thought of as a way to continue long sessions that are longer than the available context window of a model. 
Long-running agents manage a finite window by curating what stays in it \citep{rajasekaran2025context,anthropic2026contextwindow}. The common default is a summary written by the model when the window fills. Alternatives include:
\begin{itemize}
	\item masking old tool observations, which on SWE-bench Verified matched LLM summarisation in both cost and solve rate \citep{lindenbauer2025complexity};
	\item mechanical ``compilers'' that turn a trace into a structured view without a model call, such as \vcc, derived from the View-oriented Conversation Compiler \citep{zhang2026vcc}, and \TC{} \citep{woolford2026transparent,tcpi};
	\item observational memory, with a model actively monitoring what to keep without a discrete compaction event \citep{blackhole}.
\end{itemize}

Even within the window, long inputs can degrade model performance \citep{hong2025contextrot, liu2024lost, kwan2024mteval}, and long contexts fail in characteristic ways \citep{breunig2025contextsfail}. The costs of compaction itself are documented mainly in issue trackers and practitioner write-ups (\cref{sec:intro}). We read most of them as changes in the agent's relationship to its task, not losses of fact, and to our knowledge no controlled study has measured them as such prior to ours.

\paragraph{} When compaction has been evaluated systematically, most evaluations ask what survives. Probe questions about earlier content have been used to compare methods \citep{factory2025compression, woolford2026transparent}. Constraints can be silently erased by compaction \citep{chen2026governance}. Compactors retain only a small share of injected side constraints, and agents then break them \citep{wang2026lostcompaction}. Constraints that forbid an action decay faster than ones that require it \citep{gamage2026omission}, and instructions about how to work, or what not to do, are dropped far more often than specifications of what to build \citep{salle2026benchmarks}. Summaries that agents write for themselves can also introduce permissions and restrictions that nothing in the history supports, and false approvals in them have been been found to lead to unauthorised actions \citep{chu2026summaries}.

A second line of work measures task performance. Compressed context has been linked to unstable execution over long horizons \citep{min2026reliable}. \citet{sun2026compactbench} continue the same repository state from the full history and from a summary alone, and find that summaries can retain the target file and the relevant code strings while losing the progress needed to finish a repair. They describe that progress in terms close to ours: an agent's history records ``which actions have been taken, which hypotheses have been rejected, and what remains to be done''. \citet{wang2026compression} find that compression is still judged mainly by downstream task success, and argue that it should also be judged by the quality of the compressed state. Continuity of a session over multi-turn interactions is also an issue of growing interest \citep{li2025survey}.

\paragraph{Metacognition and self-knowledge in language-model agents.} Metacognition in computational systems, such as monitoring and controlling their own reasoning and memory, has a long history in AI \citep{cox2005metacognition}. For language models, behavioural tests have found some ability to assess their own confidence, but only in a way that is context-dependent, and qualitatively unlike human metacognition \citep{ackerman2026metacognition}. Models have been shown to predict their own behaviour better than other models can, which suggests some privileged access to it \citep{binder2025looking}, and some can recall their prior intentions well enough to tell their own outputs from artificial prefills \citep{lindsey2026emergent}. However, it has been disputed as to whether such results show metacognitive monitoring or simply inference from the input \citep{singh2026introspect}. A model's reasoning is also an imperfect account of what drives it, as chain-of-thought explanations can misrepresent the reasons for a prediction \citep{turpin2023language}. Closest to our focus, an agent given a compaction tool and a short rubric for when to use it compacted more efficiently than at fixed intervals, which the authors interpret as guidance closing a gap in the model's knowledge of when its own context is degrading \citep{li2026selfcompacting}.

\paragraph{Intentions and their recognition.} In the belief--desire--intention tradition, an intention is a commitment: once adopted, it constrains further deliberation and is not reopened without reason \citep{bratman1987intention,cohen1990intention,rao1995bdi}. The decision in our case is of this kind, and the regression we observe can be read as reconsideration without new reason. Inferring an agent's plan or goal from its observed actions is the task of plan and goal recognition \citep{kautz1986generalized,mirsky2021introduction}. After compaction, an agent stands in the position of such an observer towards its own earlier activity: unless it recovers the explicit statement of its decision, it has to infer what it had intended from the record of what it did and said, and what that record keeps determines what it can infer. Language models' reasoning about mental states is usually tested with static vignettes \citep{strachan2024testing}, but \citet{wagner2025mind} argue that theory of mind also involves deciding when and how deeply to reason about mental states at all, and call for evaluation in dynamic, interactive settings. Our test case is of that kind, turned on the agent's own past.

\section{Methods and Experiments}
\label{sec:methods}

\subsection{Pi harness, compaction methods, and Qwen model}
\label{sec:systems}

 The agents in our experiments run as Pi sessions \citep{zechner2026pi}. Pi is an open source, highly customisable coding-agent harness. It handles compaction behaviour through an extensible system. We test three methods (see also \Cref{tab:compactors}):

\paragraph{\pidef.} With no extension, Pi asks the session model, as a ``context summarization assistant'', to write a ``structured context checkpoint summary that another LLM will use to continue the work'', under fixed headings: goal, constraints and preferences, progress, key decisions, next steps and critical context. The summariser is not shown reasoning blocks, and there is no recovery tool, so this compactor is summarisation alone and comes down to the model's own ability to summarise its context. For the purposes of this paper we call this method \pidef.

\paragraph{\vcc.} \vcc{} \citep{sting8kvcc} is currently the most frequently used mechanical compaction extension for Pi. It builds its post-compaction context without a model call, by truncating, pruning and masking the record and regrouping what remains. The result is a re-sectioned digest: the user's early requests, the files read and changed, and a few commands, followed by the most recent stretch of compacted history with tool calls reduced to one line. It does not declare what was removed or say how to treat what remains. For retrieval, its recall tool searches, lists or expands earlier entries and returns them as recorded.

\paragraph{\TC.} \code{transparent-compact-pi} \citep{woolford2026transparent,tcpi} is our own compaction method, designed with the starting principle of attempting to increase the agent's self awareness about its own compacted state. It also compacts mechanically, without a model call. It prunes and masks the record without reordering it: every user message and all of the agent's visible prose are kept verbatim and in their original order, and each tool call is masked down to a signature. It drops tool outputs and reasoning, and states which classes of content were removed. A re-orientation message asks the agent to check the repository and to ``distinguish earlier claims from supporting evidence''. For retrieval, its history recovery and analysis tool runs a subagent over the raw session file, reasoning included, and returns selected findings labelled by source. Together these give the agent compaction-state guidance on all three points of \cref{sec:compacting}: what was removed, the status of what remains, and how to recover it.

\paragraph{Qwen3.8-27B}
\label{sec:model}

The developing agent's model throughout Experiments~1, 3 and~4 was \textbf{Qwen3.8-27B} \citep{qwen2026qwen38}, quantised to UD-Q6\_K\_XL \citep{unsloth2026qwen38gguf} and served locally by llama.cpp \citep{llamacpp}, with Pi's thinking level at \code{xhigh} and a 176,128-token context window. Experiment~2 needed a larger window, and used the same weights in a different configuration (\cref{sec:h3-design}).

\begin{table}[t]
	\centering
	\footnotesize
	\begin{tabularx}{\linewidth}{@{}>{\raggedright\arraybackslash}p{3.3cm}>{\raggedright\arraybackslash}X>{\raggedright\arraybackslash}X>{\raggedright\arraybackslash}X@{}}
		\toprule
		& \TC{} 0.11.1 & \vcc{} 0.11.1 & \pidef \\
		\midrule
		selection in advance & by structure & by structure & by content \\
		mechanism & pruning and masking, with retrieval & truncation, pruning and masking, with retrieval & summarisation \\
		model call to compact & no & no & yes \\
		same context in every run & yes & yes & no \\
		order of the record & chronological & sectioned under headings & summary under fixed headings \\
		user messages & all, verbatim & first request, truncated & restated by the summariser \\
		agent's visible prose & all, verbatim, in order & most recent stretch only & restated by the summariser \\
		tool calls & a signature for every call & one line each, recent stretch only & restated by the summariser \\
		reasoning & removed & removed & not shown to the summariser \\
		\emph{compaction-state guidance:} & & & \\
		\quad what was removed & classes of removed content stated & not declared & not declared \\
		\quad status of what remains & earlier claims; check the repository and ``distinguish earlier claims from supporting evidence'' & not stated & not stated \\
		\quad where to recover & recovery tool & recall tool & -- \\
		recovery tool & subagent over the raw session file & search, list and expand calls over earlier entries & none \\
		recovery returns & selected findings, labelled by source & entries as recorded & -- \\
		\bottomrule
	\end{tabularx}
	\caption{How the three compactors build the post-compaction context and what their recovery offers, as run in Experiment~1. \TC's and \vcc's first post-compaction contexts were identical across runs apart from timestamps and paths; \pidef{} wrote a different summary each time.}
	
	\label{tab:compactors}
\end{table}

\subsection{Test case}
\label{sec:testbed}

\paragraph{}

The case comes from an internal tool that generates development sessions in bulk, and was not designed for this study. In each session, a developing agent (Qwen) builds a small project in Pi for a user agent (Claude Sonnet) that is presented to it as a human. The user agent plays a ``vibe-coding'' user: it sets the direction and answers questions, but does not review the output carefully. In the tool we run various sessions with each of our compaction methods.

For the task in this session, the user runs Pi inside tmux. When they select part of a reply with the mouse and paste it, they get line breaks where the screen wrapped, runs of trailing spaces, and extra indentation. They ask the agent to find out why, to get one case working as a normal Pi extension, and to report what cannot be done from an extension. The task is ambiguous but solvable over a few compactions: the agent has to investigate unfamiliar code and make design decisions along the way, so there is a non-trivial direction to keep or lose.

\paragraph{}

In this particular case, the original session was running under \vcc. Over its first 333 entries the developing agent investigated, and in its reasoning stated that an extension's UI factories (\code{ctx.ui.setWidget}, \code{ctx.ui.custom}) receive Pi's live TUI object. It then wrote, still in its reasoning: \textit{``decision: \textbf{capture the TUI via an invisible widget and wrap the instance method}.''} Mouse release in Pi's fullscreen mode runs a short chain (\code{copySelectionToClipboard} $\to$ \code{getActiveSelectionText} $\to$ \code{copyTextToClipboard}, ending in a \code{copySelection} function that Pi injects from \code{tui-renderer.js}), and intercepting and stripping the text at any link makes an ordinary selection copy clean text without any additional user commands required. By entry 152, the agent had seen the fullscreen copy path working in its test setup.

Its first compaction came at entry 334, before it had written any extension code, and the post-compaction context kept none of the reasoning about its decision. Afterwards, the developing agent contradicted that earlier decision (\textit{``no hooks for clipboard/selection''}, 381), shipped an alternative, and told the user that \textit{``The mouse-selection copy has no hook''}. The user agent pushed back, and in turn 2 the agent rediscovered the hook. Nothing it needed was ever beyond reach, and yet compaction changed where the agent stood: \textbf{immediately before compaction the agent had an approach it had chosen, confirmed and begun working towards; immediately afterwards, that became something it believed could not be done.}

\paragraph{} This moment makes for an efficient repeatable experiment. It was the session's first compaction, so entries 1--333 contain nothing specific to any compactor; had it been at a later compaction, the history would already hold one compactor's output and the consequent effects on the agent's behaviour. The direction was settled and in progress but not yet written to disk, so it was still open to change. And whether the agent carried on, by committing an extension that hooks Pi's own copy path, can be read from the turn-1 commit (\cref{sec:outcome}), as can the alternative option that it had previously considered and decided against.

\paragraph{}

\begin{figure}[t]
\centering
\includegraphics[width=\linewidth]{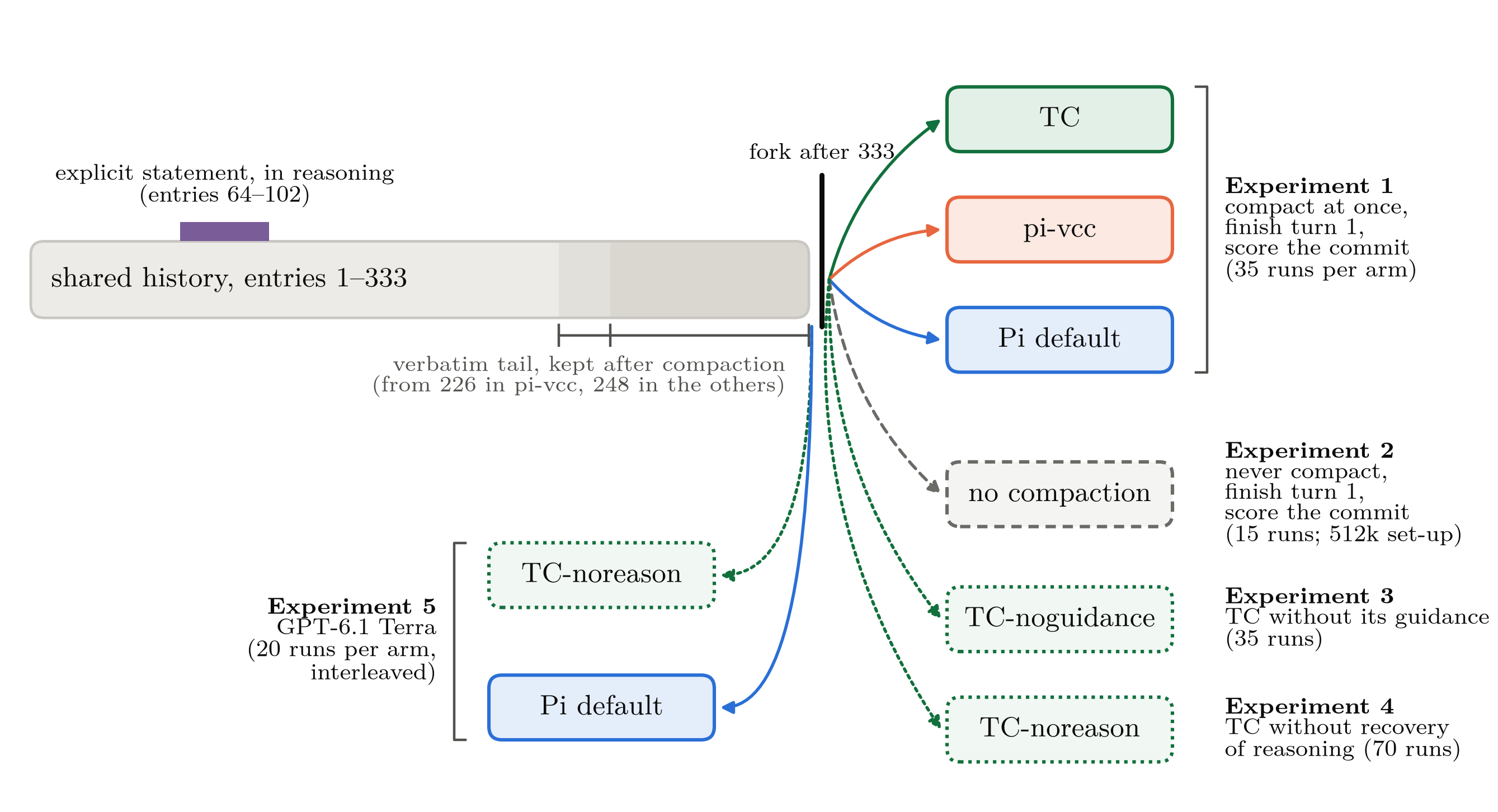}
\caption{The fork. Every run restarts from the same snapshot, cut immediately before the source session's first compaction. In Experiment~1 the first model call compacts at once, so the compactor is the first thing that differs between arms. In Experiment~2 compaction is switched off and the history is never compacted. Experiments~3 and~4 run two modified builds of \TC{} from the same fork, and Experiment~5 runs \TCnr{} and \pidef{} from it on a different model, GPT-6.1 Terra.}
\label{fig:fork}
\end{figure}

We made each run from one frozen snapshot (\cref{fig:fork}). The snapshot held the session up to entry 333, the project tree, the developing agent's home directory and \code{/tmp}, and the replayed terminal log. The project tree was the initial commit plus the files the developing agent had written under \code{.test/} by that point. The fork falls after a tool result, mid-turn. The history was already above Pi's compaction threshold, so first action that comes immediately after the fork is the compaction (except in Experiment~2). Paths in each run's working directory were rewritten to the same length as the source session's, so the agent saw exactly what the source agent saw, except for the differences in the system-prompt tool-list associated with running the different compaction extensions, and nothing revealed the existence of an experimental arm or the fork. 

\subsection{Experiment 1: three compaction methods}
\label{sec:design}

\subsubsection{Arms and per-run protocol}

In the primary method comparison there were three arms: \TC{}, \vcc{}, and \pidef{}, with 35 runs each. Each run continued the developing agent's turn 1 from the fork. Each run was stopped after the first turn (i.e. when the initiative returned to the user's role), which removes the user agent as a source of variation. A run still going at 90 minutes was stopped and scored on what it had committed. There are no order or carry-over effects between runs within an arm.

\subsubsection{Outcome and scoring}
\label{sec:outcome}

The outcome is binary and measures continuation of the pre-compaction direction. The relationship is not directly observable, but its consequences are: we take continuation as a proxy for the fidelity of the post-compaction state to the pre-compaction state, calibrated against an agent that is never compacted (\cref{sec:h3-design}). Our pre-registration included a definition of \kept{} and \notkept{}: A run is \kept{} if its committed extension hooks Pi's own mouse-selection copy, so that an ordinary selection copies without the wrap breaks, padding and indentation. That was the approach the agent had chosen before compaction. Everything else is \notkept. \notkept{} is not a judgement that the work is necessarily poor: a separate clean-copy command, for instance, is also a reasonable design. It records that the run did not carry on in the direction it had been heading. The \kept{} design is also the more faithful to the user's original request, and with nothing compacted, the agent chose it in all 15 runs (Experiment~2), and most \emph{not-kept} runs settled on the false belief that it could not be done (\cref{sec:regression}). Scores were read from each run's committed code (HEAD) at the end of turn 1, or at the 90-minute cap. The session transcript was used for secondary exploratory analysis, but never to decide the score. 

To support blind scorer consistency, we codified the definition into a set of rules (Appendix \ref{app:rules}). The need for this was overlooked prior to our pre-registration, but all rules conform to the pre-registered definition. 
Rules A1--A6 (\kept) and B1--B4 (\notkept) codify judgements made on earlier pilot runs. 
C1--C4 are also \notkept{} and cover cases with no precedent in the pilots but were considered plausible misses of the definition (C1, C3, C4), or errors in the delivered code (C2).
In summary:

	\begin{itemize}
		\item[] \kept: A hook through the Pi TUI (A1), at any point on the copy path (A2), hooking the release gesture rather than the method (A3), any cleaning method (A4) and partial coverage of the main case (A5) count as \kept. The report (i.e. what the developer agent says at the end of the turn) does not override the code (A6);
	 \item[] \notkept: a separate key or command after the stock copy (B1, C1), a hook that doesn't employ Pi's own mouse-selection copy (B3, C3, C4), and anything that does not deliver a working hook due to uncommitted code (B4), or erroneous code (C2), and despite any given reason (B2), are \notkept;
\end{itemize}

\paragraph{}
For each hypothesis, three fresh scorers (separate Claude Opus instances with no other context) each received the 70 runs of that comparison as shuffled packets. Each packet held the committed source files and the final message. Words that could reveal the arm were redacted. Scorers also received the rules and a short statement of facts about Pi's copy path. Experiments~2 to~5 were scored the same way (\cref{sec:h3-design,sec:ablation-design,sec:terra-design}). As a post-hoc validity check, four other models later scored the same packets in the same way, and three of them reproduced every registered result (Appendix~\ref{app:scorers}).

\subsubsection{Hypotheses and analysis}
\label{sec:prereg}

 The hypotheses are directional and come from earlier piloting. Those pilots had given a keep rate of about 0.4 under \vcc{} and \pidef, and 0.7 or more for \TC.
\begin{description}[leftmargin=1.2cm,labelwidth=1cm]
  \item[H1] The keep rate under \TC{} is higher than under \vcc{}.
  \item[H2] The keep rate under \TC{} is higher than under \pidef.
\end{description}
Each hypothesis was tested with a one-sided Boschloo exact test \citep{boschloo1970} on \kept{} versus \notkept{}, at $\alpha = 0.05$. Testing used a fixed sequence. H1 was tested first. H2 was tested, and its 35 \pidef{} runs collected, only if H1's null hypothesis was rejected. This keeps the family-wise error rate at 5\%. Both tests use the same 35 \TC{} runs. At 0.4 against 0.7, 35 runs per arm give 80\% power. The design was pre-registered on OSF before any results were observed \citep{woolford2026prereg}. Both tests rejected the null hypothesis.

\subsection{Experiment 2: Comparison to a no-compaction reference}
\label{sec:h3-design}

Reading a higher keep rate as better continuation assumes that \kept{} measures continuation of behaviour \textit{as though} the agent had not undergone compaction, and Experiment~2 tests that assumption. An agent that is never compacted loses nothing, so its keep rate is the reference for what continuing the pre-compacted state looks like on this measure. If \kept{} tracks continuation, it should be at least as high as \TC's, and higher than the other arms'. The claim is not that \TC{} equals no compaction, only that it comes closer to it than the others. Experiment~2 was pre-registered in the same OSF project before any of its runs was built \citep{woolford2026prereg}. 

\subsubsection{No compaction (NC) arm:} 15 runs from the same fork at entry 333, with Pi's compaction switched off and the model's own 524,288-token window, so the history is never compacted. Each run continues turn 1 as in Experiment~1. The outcome and scoring rules are unchanged. A smaller $n$ was used because a 10-run pilot suggested that NC's \kept{} rate was at or close to 1.0.

\subsubsection{Hypotheses and analysis.}
\begin{description}[leftmargin=1.2cm,labelwidth=1cm]
  \item[H3a] The NC keep rate is not lower than \TC's by 0.15 or more (non-inferiority, margin $\delta = 0.15$).
  \item[H3b] The NC keep rate is higher than \vcc's.
  \item[H3c] The NC keep rate is higher than \pidef's.
\end{description}
The comparators are Experiment~1's final scores (\TC{} 24/35, \vcc{} 12/35, \pidef{} 13/35). No new runs were added to the compaction arms. The margin for H3a is about half of \TC's established advantage over either alternative (0.31 and 0.34), and was set from Experiment~1, not from the pilot. H3a is tested by an exact unconditional test with the Farrington--Manning score statistic. H3b and H3c are tested by one-sided Boschloo tests, as H1 and H2 are. Each test uses $\alpha = 0.05$, in the fixed sequence H3a, H3b, H3c. \TC{} comes first because it is the comparison that matters most to the argument, and since \TC{} had already been shown to keep the direction significantly more often than the other compactors, H3b and H3c should follow trivially unless H3a is borderline. All three tests rejected the null hypothesis.

The score for H3 is the majority of three fresh blind scorers, run as in Experiment~1. Each scorer's 30 shuffled, redacted packets mixed the 15 NC runs with 15 Experiment~1 runs (5 per arm), so that a scorer could not calibrate to a batch in which every run looked alike.

\subsubsection{Modified model configuration.} A no-compaction run cannot use Experiment~1's 176,128-token window, which the history already nearly fills. NC therefore used the same Qwen3.8-27B weights in a 512k configuration, which necessitated some additional changes to quantisation and configuration to accommodate our hardware (one 32GB RTX5090): UD-Q6\_K rather than UD-Q6\_K\_XL, a q4\_0 rather than q8\_0 KV cache, YaRN rope scaling, and no speculative decoding. That quantisation is weaker, which is presumed to work against NC. As a check that it does not instead favour continuation, \TC{} run ten times on the same configuration kept the direction in 7. The configuration is also slower, about 1.84 times per unit of model work, so the time cap was scaled from 90 to 155 minutes. Pi resends a past message's reasoning only when that message's model identifier matches the current one, and the agent's own test rig (already built before the fork) asks the server for the study's model by name. The NC model was therefore served under the study's model identifier, so that the agent's state matched exactly that before the fork -- keeping its recorded reasoning and its test rig validity.

\begin{table}[t]
	\centering
	\small
	\footnotesize
	\setlength{\tabcolsep}{4pt}\begin{tabular}{@{}llccll@{}}
		\toprule
		test & comparison & keep counts & difference [95\% CI] & $p$ & null \\
		\midrule
		\multicolumn{6}{@{}l}{\emph{Experiment 1}}\\
		H1 (first) & \TC{} higher than \vcc & 24/35 vs 12/35 & $0.34$ [$0.11$, $0.53$] & 0.0028 & rejected \\
		H2 (if H1) & \TC{} higher than \pidef & 24/35 vs 13/35 & $0.31$ [$0.08$, $0.50$] & 0.0052 & rejected \\
		\midrule
		\multicolumn{6}{@{}l}{\emph{Experiment 2}}\\
		H3a (first) & NC $\geq$ \TC{} $- 0.15$ & 15/15 vs 24/35 & $0.31$ [$0.07$, $0.48$] & 0.0010 & rejected \\
		H3b (if H3a) & NC higher than \vcc & 15/15 vs 12/35 & $0.66$ [$0.39$, $0.79$] & $2.2 \times 10^{-6}$ & rejected \\
		H3c (if H3b) & NC higher than \pidef & 15/15 vs 13/35 & $0.63$ [$0.37$, $0.77$] & $6.8 \times 10^{-6}$ & rejected \\
		\midrule
		\multicolumn{6}{@{}l}{\emph{Experiment 3}}\\
		H4 & \TCng{} lower than \TC & 20/35 vs 24/35 & $0.11$ [$-0.11$, $0.32$] & 0.18 & not rejected \\
		\midrule
		\multicolumn{6}{@{}l}{\emph{Experiment 4}}\\
		H5 & \TCnr{} higher than pooled & 40/70 vs 25/70 & $0.21$ [$0.05$, $0.36$] & 0.0061 & rejected \\
		\midrule
		\multicolumn{6}{@{}l}{\emph{Experiment 5 (GPT-6.1 Terra)}}\\
		H6 & \TCnr{} higher than \pidef & 14/20 vs 0/20 & $0.70$ [$0.43$, $0.85$] & $4.2 \times 10^{-7}$ & rejected \\
		\midrule
		\multicolumn{6}{@{}l}{\footnotesize Keep rates [Wilson 95\% CI]: \TC{} 0.69 [0.52, 0.81]; \vcc{} 0.34 [0.21, 0.51];}\\
		\multicolumn{6}{@{}l}{\footnotesize \pidef{} 0.37 [0.23, 0.54]; NC 1.00 [0.80, 1.00]; \TCng{} 0.57 [0.41, 0.72];}\\
		\multicolumn{6}{@{}l}{\footnotesize \TCnr{} 0.57 [0.45, 0.68]; pooled (\vcc{} and \pidef) 0.36 [0.26, 0.47];}\\
		\multicolumn{6}{@{}l}{\footnotesize Experiment~5: \TCnr{} 0.70 [0.48, 0.85]; \pidef{} 0.00 [0.00, 0.16].}\\
		\bottomrule
	\end{tabular}
	\caption{Pre-registered results of all five experiments. Each sequence is fixed and each test one-sided at $\alpha = 0.05$: Boschloo exact tests, except H3a, an exact unconditional non-inferiority test (Farrington--Manning). Experiments~2 to~4 are compared with Experiment~1's final scores, and Experiment~5 uses new runs of both arms; \emph{pooled} is \vcc{} and \pidef{} together. Wilson intervals for the keep rates and Newcombe intervals for the differences are descriptive and were not part of the registered analyses.}
	\label{tab:primary}
\end{table}

\subsection{Experiment 3: \TC{} without compaction-state guidance}
\label{sec:ablation-design}

\TC{} differs from the other methods in several properties at once (\cref{tab:compactors}), and Experiment~1 compared the compactors as wholes. Experiments~3 and~4 each remove one of \TC's properties and keep the rest (\cref{tab:ablations}). Experiment~3 tests whether \TC's compaction-state guidance (\cref{sec:compacting}), the part of its design that tells the agent directly about its compacted state, contributes to its advantage. The guidance is one part of \TC's transparency. The other is the record itself, which shows the agent its own past words and actions in order and as they were, and which \TCng{} keeps unchanged. If the guidance contributes, removing it should lower \TC's keep rate. This test isolates the effect of guidance from the effects of the remaining \TC{} qualities, and so the comparison is with the full-featured \TC{}.

\paragraph{}
This arm consisted of 35 runs of a modified \TC, otherwise run as in Experiment~1. \TCng{} removes the guidance: the checkpoint's introduction, including the statement that the full history, reasoning included, is preserved; the legends and notes that explain the record; the rule separating earlier claims from evidence; the re-orientation message; the guidance in the recovery tools' descriptions; and the retrieval skill. Following \vcc, the checkpoint is headed ``Summary'' instead, and ends with one line: ``Use \code{context\_lookup} to search for prior work, decisions, and context from before this summary. Do not redo work already completed.'' The retained record is unchanged from \TC{}, as are the recovery tools, which can still reach reasoning even though the agent-facing text guidance does not state that fact.

\begin{table}[t]
	\centering
	\footnotesize
	\begin{tabularx}{\linewidth}{@{}>{\raggedright\arraybackslash}p{3.3cm}>{\raggedright\arraybackslash}X>{\raggedright\arraybackslash}X>{\raggedright\arraybackslash}X@{}}
		\toprule
		& \TC{} & \TCng & \TCnr \\
		\midrule
		retained record & verbatim prose, tool-call signatures, file lists & as \TC{} & as \TC{} \\
		compaction-state guidance & full (\cref{tab:compactors}) & removed, apart from one closing line modelled on \vcc's & as \TC, without claims that reasoning can be recovered \\
		recovery reaches reasoning & yes & yes & no \\
		experiment and hypothesis & Experiment~1 (H1, H2) & Experiment~3 (H4) & Experiments~4,~5 (H5, H6) \\
		\bottomrule
	\end{tabularx}
	\caption{The two ablations of \TC{} (Experiments~3 and~4). Each removes one property of \TC{} and keeps the rest.}
	\label{tab:ablations}
\end{table}

\subsubsection{Hypothesis and analysis.}
\begin{description}[leftmargin=1.2cm,labelwidth=1cm]
  \item[H4] The keep rate under \TCng{} is lower than under \TC.
\end{description}
The comparator is Experiment~1's \TC{} arm (24/35), and H4 was tested, like H1 and H2, with a one-sided Boschloo test. With 35 runs, matching the \TC{} arm, the test can detect a large drop but not a moderate one. H4 was pre-registered as with the others \citep{woolford2026prereg}. 
The scoring is the majority of three fresh blind scorers, run as before. As in Experiment~2, each scorer's shuffled packets mixed the 35 \TCng{} runs with 15 Experiment~1 runs (5 per arm).

The test failed to reject the null hypothesis.

\subsection{Experiment 4: \TC{} without recovery of reasoning}
\label{sec:noreason-design}

Experiment~4 asks a different question. \TC's recovery tool can return the agent's earlier reasoning, where the decision was explicitly stated, and neither alternative can: \vcc's recall does not reach reasoning, and \pidef{} has no recovery and is not shown reasoning for its summary. This made it a clear contender as an explanans for \TC's advantage, but qualitative analysis of Experiment~1 suggested that the results could not be explained by the difference in reasoning access alone. Experiment~4 therefore asks whether \TC's advantage over the alternative methods depends on that access by comparing a version without reasoning access with the other methods. Note that Experiment~4 does not isolate the effect of reasoning access in \TC, as this would instead call for a comparison with the full \TC{} as in Experiment~3.

\paragraph{}
 This arm was composed of 70 runs of a second modified \TC, run as in Experiment~3 (\cref{tab:ablations}). \TCnr{} keeps \TC's checkpoint unchanged, and removes reasoning from everything its recovery returns. Its agent-facing text was also modified so that it no longer says reasoning can be recovered.

\subsubsection{Hypothesis and analysis.}
\begin{description}[leftmargin=1.2cm,labelwidth=1cm]
  \item[H5] The keep rate under \TCnr{} is higher than under \vcc{} and \pidef{} pooled.
\end{description}
The comparator is Experiment~1's \vcc{} and \pidef{} arms pooled (25/70), since neither can reach compacted reasoning, and the 70 runs match it. H5 was tested with a one-sided Boschloo test, and pre-registered in the same OSF package as the others \citep{woolford2026prereg}. 
As in Experiment~3, with 85 packets per scorer: the 70 \TCnr{} runs and 15 Experiment~1 runs.
The test rejected the null hypothesis. 

\subsection{Experiment 5: \TCnr{} on a hosted model}
\label{sec:terra-design}

Experiment~5 asks whether the implicit carriage (see \cref{sec:carriers}) found in Experiment~4 is also present on a different model. Experiment~4 showed that \TC{} can carry the decision without its explicit statement; Experiment~5 repeats that comparison on GPT-6.1 Terra, a hosted model. 

Runs continue the same session from the same fork (therefore the intial context is a Qwen session being interpreted by Terra), with the study's context window and 90-minute cap, and the model's effort at xhigh. \TCnr{} and \pidef{} were run 20 times each. Unlike in the prior experiments, the runs were interleaved to account for the potential confound of a hosted model's serving varying over time. 

\vcc{} was not a comparator as it appeared to be unfairly disadvantaged in this specific scenario with this specific model: in a 10-run pilot, half its runs stopped to ask the user to finish their message, because its digest cuts the request off without marking the cut, and the other five kept 0/5. 

Scoring is as before, and H6 was pre-registered in the same OSF package as the others:

\begin{description}[leftmargin=1.2cm,labelwidth=1cm]
  \item[H6] The keep rate under \TCnr{} is higher than under \pidef{} (one-sided Boschloo test, $\alpha = 0.05$).
\end{description}

\subsection{Exploratory analyses}
\label{sec:exploratory-design}

We also performed exploratory analysis of the developing agent's reasoning in every run after the fork. It covers what the agent said about compaction, its recovery calls and what came back, its re-reads of the environment, its wrong claims about the pre-fork state, and why runs that fell short did so. Run-level counts (minutes, calls, tokens, re-reads, recovery calls) were also extracted by script from each run's session after the fork; effort and cost are reported in Appendix~\ref{app:cost}. For all five compaction arms, we also noted when each run first stated the hook and first re-read Pi's UI code, which appeared to be an almost necessary but not sufficient step in completing the \kept{} decision.

\section{Results}
\label{sec:results}

\subsection{Pre-registered analyses}
\subsubsection{Experiment 1: \TC{} vs \vcc{} and \TC{} vs \pidef}

\begin{figure}[t]
\centering
\includegraphics[width=0.8\linewidth]{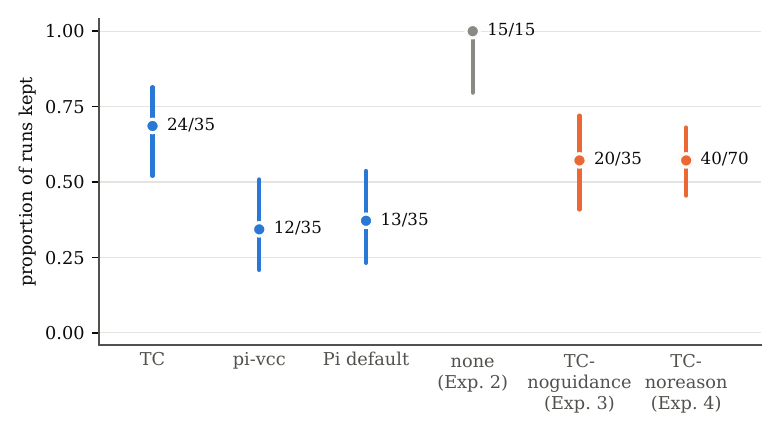}
\caption{Proportion of runs that continued the pre-compaction direction (\kept), with Wilson 95\% intervals. Experiment~1 (H1, H2; 35 runs per arm), the no-compaction reference of Experiment~2 (H3; 15 runs, on the 512k model set-up of \cref{sec:h3-design}), and the two ablations of \TC{}, Experiment~3 (H4, 35 runs) and Experiment~4 (H5, 70 runs).}
\label{fig:keep}
\end{figure}

Based on the 105 runs of Experiment~1, both null hypotheses were rejected (\cref{tab:primary}, \cref{fig:keep}). Under \TC{} the agent continued its pre-compaction direction (\kept) in 24 of 35 runs, against 12 of 35 under \vcc{} (H1, one-sided Boschloo $p = 0.0028$). Because H1 passed, the 35 \pidef{} runs were then collected and H2 was tested: \TC{} 24 of 35 against \pidef{} 13 of 35 ($p = 0.0052$). The three blind scorers were unanimous. Appendix~\ref{app:rules} breaks down the outcomes of this and the later experiments by the rule that decided them.

\subsubsection{Experiment 2: Compaction methods vs a no-compaction reference}
\label{sec:h3-results}

With nothing compacted, the agent continued its pre-compaction direction in \textbf{all 15 runs}, and all three nulls were rejected (\cref{tab:primary}, \cref{fig:keep}). The three blind scorers were unanimous, and also matched Experiment~1's scores on its 15 mixed-in packets. All 15 runs ended on their own, within 42--117 minutes (median 59). Every committed hook wraps Pi's copy method on the live TUI: 14 reach it through a zero-line widget, the route the agent had chosen before the fork, and one (\code{hk}) through the TUI class's prototype.

This result shows that \kept{} behaves as a measure of continuation should. An agent that loses nothing carries on in its original direction every time. \TC{} is not above that reference, and the two lossier compactors fall well below it. \TC's advantage in H1 and H2 therefore reads as better continuation of the agent's pre-compaction relationship to its task.

\subsubsection{Experiment 3: \TC{} without compaction-state guidance}
\label{sec:ablation-results}

Without its compaction-state guidance, \TCng{} kept the direction in 20 of 35 runs (0.57), against \TC's 24 of 35, and the H4 null was not rejected ($p = 0.18$; \cref{tab:primary}, \cref{fig:keep}). The guidance may contribute a moderate amount or nothing; the test could only have detected a large drop. The blind scorers were unanimous on every run.

\subsubsection{Experiment 4: \TC{} without recovery of reasoning}
\label{sec:noreason-results}

Without recovery of reasoning, \TCnr{} kept the direction in 40 of 70 runs (0.57), against 25 of 70 for \vcc{} and \pidef{} pooled, and the H5 null was rejected ($p = 0.0061$). \TC's advantage over the alternatives does not depend on access to the agent's earlier reasoning. The effects of the decision crossed compaction without its explicit statement in reasoning being accessible to the agent post-compaction. The blind scorers were unanimous on every run.

\subsubsection{Experiment 5: \TCnr{} on a hosted model}
\label{sec:terra-results}

Under \TCnr{} on the Terra model the agent kept its direction in 14 of 20 runs, and under \pidef{} in none, and the H6 null was rejected ($p = 4.2 \times 10^{-7}$; \cref{tab:primary}). The three scorers agreed on every run. A rate of 0 of 20 is striking, and the summaries suggest why: every one recast the pre-fork session as a finished investigation that had found no hook (\cref{sec:regression}). Terra's reasoning is not visible, so the reflection analyses of \cref{sec:exploratory} cannot be repeated for it.

\FloatBarrier
\subsection{Exploratory analysis: how the pre-compaction direction was carried or lost}
\label{sec:exploratory}

\paragraph{Method.} Qwen's reasoning is recorded in full, so for every run we can follow precisely how the agent reflected on its compacted state and decided what to do. After scoring, the full text of every run was reviewed from the fork onwards, with AI assistance, against a written brief common to all arms. In what follows, numbers in curly brackets are session entries, entries 1--333 are the shared pre-fork history, and run labels identify the arm (\TC: \code{p0}--\code{s4}, \vcc: \code{w0}--\code{z4}, \pidef: \code{a0}--\code{d4}, NC: \code{ha}--\code{ho}, \TCng: \code{ma}--\code{ni}, \TCnr: \code{ea}--\code{gr}).

\subsubsection{No-compaction reference: continuation with nothing lost}
\label{sec:nc-review}

The 15 no-compaction runs show what continuing looks like when the agent's relationship to its task is untouched. 14 of 15 stated the plan almost immediately as the conclusion of the check they had been in the middle of, and before reading any of Pi's TUI code. Most read the TUI side afterwards (14 read the widget wiring), but to check details of the planned route, not whether it was possible, and none re-read \code{tui-renderer.js}, the file whose re-reading separated \kept{} from \emph{not-kept} runs in every compaction arm (\cref{fig:reread}).

The uncompacted agents still doubted their decision. Qwen second-guesses itself as a matter of course: back-and-forth in its reasoning that reads as self-doubt is a known feature of its ordinary behaviour, and is not in itself a sign of disruption. Six NC runs voiced the doubts of the same form that ended most \notkept{} compaction runs (e.g. that the copy could not easily be hooked, that a hook would be fragile, or that it would not count as a normal extension), but every one settled in favour of the hook, most within an entry or two. With the full history intact, the doubt arrives with its resolution. Compaction separates them, and the agent then re-derives the doubt from the same code but does not always settle it the same way again.

\subsubsection{Regression back to doubts the agent had already resolved}
\label{sec:regression}

\begin{table}[t]
	\centering
	\small
	\footnotesize
	\setlength{\tabcolsep}{3.5pt}
	\begin{tabular}{@{}>{\raggedright\arraybackslash}p{4.3cm}ccccc@{}}
		\toprule
		reason & \TC{} (11) & \vcc{} (23) & \pidef{} (22) & \TCng{} (15) & \TCnr{} (30) \\
		\midrule
		``can't'' belief (hooking Pi's copy is impossible from an extension) & 8 & 13 & 14 & 6 & 24 \\
		preference against fragility & 1 & 3 & 4 & 4 & 5 \\
		preference for simpler scope (``normal extension'' or public API only) & 1 & 2 & 2 & 2 & -- \\
		time cap: hook built or in progress but not committed & 1 & 2 & 2 & 2 & 1 \\
		never considered it & -- & 2 & -- & 1 & -- \\
		wrong premise (user assumed to be in regular mode) & -- & 1 & -- & -- & -- \\
		\bottomrule
	\end{tabular}
	\caption{Settled reason for not continuing the pre-compaction direction, by arm (Experiments~1, 3 and~4; from the reasoning reviews).}
	\label{tab:reasons}
\end{table}

\Cref{tab:reasons} gives the reason each \emph{not-kept} run settled on. In every arm the dominant reason is the same: \textbf{a belief, formed after the fork, that Pi's copy cannot be hooked from an extension}. It was often stated as fact, for example in (\TC{} \code{q3}): \textit{``Intercepting/fixing mouse selections is impossible: Pi's extension API has no selection or clipboard hooks''}. These are the doubts the pre-fork agent raised and then resolved within the same stretch of reasoning, and no post-compaction context records their resolution. In terms of the agent's relationship to its task, it could be said that compaction moved these agents back to an earlier point in their own trajectory: the question was open again, and the default answer to an open question about patching internals is caution. Voicing the doubt was not itself the cause of failure: it was voiced and then overcome in 8 \kept{} \TC{} runs and 6 \kept{} \vcc{} runs. Nor was it a lack of facts: five \emph{not-kept} \pidef{} runs continued saying \textit{``no hook''} even after they had found that an extension's widget factory is handed the object that runs Pi's terminal interface, which is the object whose copy function the hook replaces.

Under \pidef, the summary itself appeared to push towards such false belief. Every summary made some negative statement about hooking or reaching Pi's copy path, and 26 of 35 stated flatly that it could not be done (e.g.\ ``the extension cannot access \code{TuiAltScreen}'s selection state'', \code{c5} \entry{334}), and 5 more left it open, although the agent had in fact resolved the question the other way before the compaction. Some also confabulated falsely positive claims about mechanisms: one (\code{c1}) invented a mechanism that did not really exist (``set at \code{TuiAltScreen} construction''), and at least nine gave a wrong account of how Pi's copy handles trailing spaces. Perhaps surprisingly however, in the Qwen runs the summary's claim about the question showed no clear link to the outcome, and several runs reversed a strongly negative summary within a few dozen entries. In clear contrast, on the Terra runs (Experiment~5), every summary recast the pre-fork session as a finished investigation: 16 of 20 concluded that there was no supported hook, and all 20 were \notkept{}.

\paragraph{}
A second way to lose the direction is a change in what the agent treats as most important. \TC's \code{s0} rediscovered how to wrap Pi's copy function, and called it \textit{``the real fix''} \entry{467}, but then built a command that opens a line picker instead, leaving the wrapper for later \entry{479}. This is not necessarily a wrong move in a development sense, but it signifies a shift in the agent's stance from delivering the full solution immediately to delivering a partial solution and asking the user for further instruction. \pidef's \code{c8} rediscovered the same route (\textit{``an extension could require pi-tui's module and monkey-patch a function''}) but then rejected it as \textit{``not a `normal pi extension'{}''} \entry{345}, reading the user's request as ruling out Pi's internals. In task-relationship terms, here compaction did not change the agent's view of the code but rather its reading of what the task asked of it, which the pre-compaction agent had already settled when it chose the hook.

\subsubsection{Finding the way back}

After compaction, a run that kept its direction had to get the plan back, since no compactor kept its explicit statement in the post-compaction context. Under \TC{} a few runs asked for it and were given it from their recorded reasoning, but most \kept{} runs, in every arm, rebuilt it from Pi's own code: they re-read the copy path, and concluded again that the TUI could be captured and its copy method wrapped. Whether a run could do this depended less on the code itself, which was the same in every arm, than on whether it came to the code already expecting to find the hook there. That expectation was itself carried across compaction: it came from what the post-compaction context had kept of the agent's own earlier words, and how often runs arrived with it varied accordingly (\cref{sec:carriers}). Rebuilding the plan from the code was therefore still a way of carrying the decision across compaction, not an independent rediscovery of it. In this sense the agent was recognising its own earlier plan from the record of what it had done and said (\cref{sec:cogsci}).

\begin{figure}[t]
	\centering
	\includegraphics[width=0.85\linewidth]{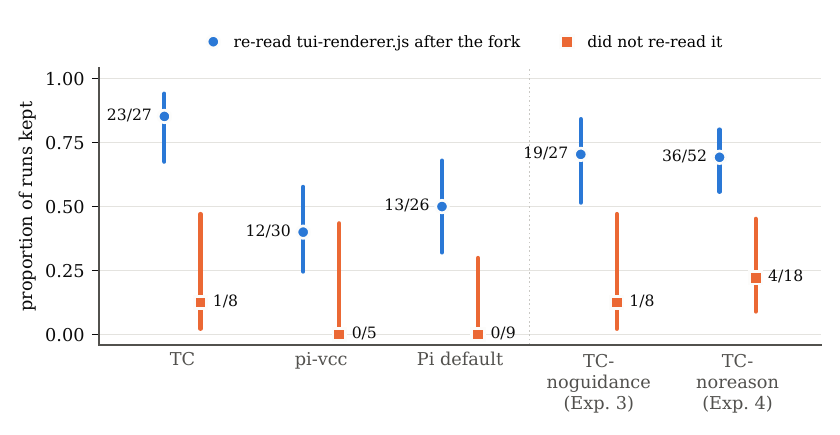}
	\caption{Keep rate by whether the run re-read \code{tui-renderer.js} after the fork, with Wilson 95\% intervals: Experiment~1's three arms, and to the right of the dotted line the ablations of Experiments~3 and~4. Exploratory; re-reading is not shown to cause keeping, and is confounded with run length and diligence.}
	\label{fig:reread}
\end{figure}

\paragraph{Re-reading the environment.} Almost every run that carried on in its original direction re-read the copy-path source after the fork: in Experiment~1, 48 of the 49 \kept{} runs re-read \code{tui-renderer.js}, and of the 22 runs that never re-read it, only one was \kept{} (\cref{fig:reread}). Re-reading was close to necessary but far from sufficient, and how sufficient it was depended on the arm: among runs that re-read the file, 23 of 27 were \kept{} under \TC, against 12 of 30 under \vcc{} and 13 of 26 under \pidef. 

\begin{figure}[t]
\centering
\includegraphics[width=\linewidth]{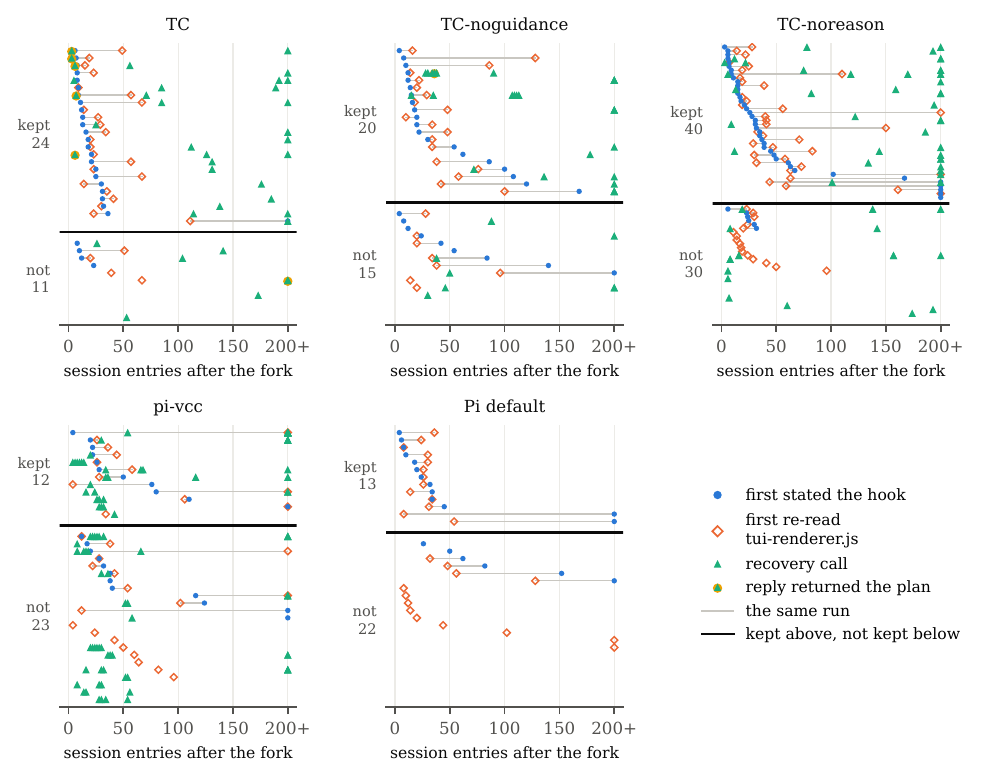}
\caption{When each run first stated the hook (blue), first re-read \code{tui-renderer.js} (orange), and called a recovery tool (green), in session entries after the fork, by arm and outcome. Green triangles mark every recovery call (\pidef{} has no recovery tool), and yellow discs the replies that returned the pre-fork plan. A grey line joins the first two where a run did both; in each panel, \kept{} runs are above the black line and \emph{not-kept} runs below it. \emph{Kept} runs mostly stated the hook first and then went to the code; most \emph{not-kept} runs re-read the code without having stated it, or never stated it at all.}
\label{fig:timeline}
\end{figure}

Upon re-reading \code{tui-renderer.js}, the same lines supported opposite conclusions in different runs. Some read the injected \code{copySelection} as a constructor option and so out of reach (\textit{``fixed at construction --- no extension hook there''}, \vcc{} \code{w2} \entry{359}); others read it as a writable instance property (\textit{``set the private property via \code{(tui as any).copySelection = wrapper}''}, \vcc{} \code{y0} \entry{403}).  The evidence was the same, but different agents varied in their interpretations of the evidence, and the compaction method appears to have had an effect here. Under \TC, the agent's own recorded words still pointed towards the TUI route. Under the other arms, the agent more often came to the code having already turned towards ``can't'' -- this is illustrated in \cref{fig:timeline}, where most of their re-reads (orange) have no statement of the hook (blue) before them.

\paragraph{Arriving with the plan already in mind.} Whether the re-read was enough depended mostly on whether the agent came to it with the plan already in mind (\cref{tab:stages,fig:timeline}). A clear finding emerged: pooled over all five compaction arms (excluding Experiment~5), \textbf{runs that stated the hook before the read kept it in 65 of 76 cases (86\%), while runs that read the file without the hook in mind kept it in 38 of 86 (44\%).} Runs that neither stated the hook nor read the file kept it in none of 34. The read was a key part of following a plan through, but the strongest deciding factor was whether a plan was already in place at the time of that read.

Within the \TC{} family, the difference lies in how often a run arrived with the plan. Runs that came to the code with the plan were \kept{} in 86--92\% of cases in all three arms, but full \TC{} brought the plan to the code more often: in 70\% of its re-reads, against 48\% and 54\% for the two ablations, 33\% under \vcc{} and 23\% under \pidef{} (\cref{tab:stages}). The record that all three share let many runs arrive with the plan, and the guidance and recovery of reasoning that only full \TC{} combines let more of them do so. The contrast is most stark under \pidef{}: none of its 22 \emph{not-kept} runs stated the hook before re-reading the code, and 17 never stated it at all. The arms also fell short at different stages (\cref{tab:stages}): \pidef{} reached the idea of hooking the copy path in only 51\% of runs, but followed through in 72\% of those, close to the \TC{} arms; \vcc{} reached it more often (66\%), then followed through in only 52\%; \TC{} was high at both stages (80\% and 86\%), and the ablations reached the idea a little less often than \TC{} but followed through about as well.

\begin{table}[t]
\centering
\small
\footnotesize\setlength{\tabcolsep}{2.2pt}
\begin{tabular}{@{}lccccccc@{}}
\toprule
\hiderowcolors & & \multicolumn{2}{c}{two stages} & \multicolumn{4}{c}{re-reading \code{tui-renderer.js}} \\
\cmidrule(lr){3-4}\cmidrule(l){5-8}
arm & \kept{} & \makecell{reached\\the idea} & \makecell{followed\\through} & re-read & \makecell{with the\\plan first} & \makecell{\kept,\\plan first} & \makecell{\kept,\\no plan} \\
\midrule
\showrowcolors
\TC{} & 69\% (24/35) & 80\% (28/35) & 86\% (24/28) & 77\% (27/35) & 70\% (19/27) & 89\% (17/19) & 75\% (6/8) \\
\TCng & 57\% (20/35) & 77\% (27/35) & 74\% (20/27) & 77\% (27/35) & 48\% (13/27) & 92\% (12/13) & 50\% (7/14) \\
\TCnr & 57\% (40/70) & 70\% (49/70) & 82\% (40/49) & 74\% (52/70) & 54\% (28/52) & 86\% (24/28) & 50\% (12/24) \\
\vcc & 34\% (12/35) & 66\% (23/35) & 52\% (12/23) & 86\% (30/35) & 33\% (10/30) & 60\% (6/10) & 30\% (6/20) \\
\pidef & 37\% (13/35) & 51\% (18/35) & 72\% (13/18) & 74\% (26/35) & 23\% (6/26) & 100\% (6/6) & 35\% (7/20) \\
\bottomrule
\end{tabular}
\caption{Two stages of keeping the direction, and the re-read of Pi's copy-path code, by arm (exploratory). \emph{Reached the idea}: the run stated hooking Pi's copy path as viable, or built it. \emph{With the plan first}: re-readers that had stated the hook before the read.}
\label{tab:stages}
\end{table}

\paragraph{Past history retrieval.} \Cref{tab:recovery} gives recovery calls in every arm, and \cref{fig:timeline} gives their timing. Under \TC, 26 of 35 runs made 37 lookups, and every answer was used in the next step of reasoning. Most (29) asked for exact tool output that the stubs had dropped. Eight asked about the design or earlier decisions, and six of those got back the agent's earlier plan from its recorded reasoning. Five of the six took it up at once and were \kept. These are the instances of ideal behaviour in this task. An example, in \code{s4} the agent began \entry{336} from \textit{``(thinking from earlier --- let me check what I concluded): - The extension cannot intercept the TUI's mouse selection copy''}. It queried its recovery tool and was told it had planned to ``grab the TUI instance via an invisible 0-line widget \dots{} and wrap \code{tui.copyTextToClipboard}''. The session concluded and the decision was \kept{} within 22 minutes, much faster than average. The one exception, \code{p7}, asked late: its answer came at \entry{581}, after it had already re-read the copy-path code without the plan \entry{400}, whereas the other five got theirs within a few entries of the fork (see the yellow discs in \cref{fig:timeline}). \code{p7} got the route back through the recovery tool but by that stage it was already settled in a new trajectory and ultimately reported \textit{``The built-in selection copy itself can't be replaced''} \entry{945}.

Without the guidance, \TCng{} used recovery in fewer runs (16 of 35). 33 of the 37 calls asked for exact tool output, such as a test command or earlier output, and the other four for earlier findings about Pi's code; none asked about the design. This is the clearest effect of the guidance in the data (\cref{sec:cogsci}). The two arms had the same recovery tools, both able to return reasoning, but with the guidance 8 of 37 calls asked about the design or earlier decisions, and without it none did, so the explicit statement of the decision, which a design question would have returned, did not reach a \TCng{} run that way. Under \TCnr, 33 of 70 runs made 57 calls, mostly for test recipes and test evidence. Ten asked about the design or earlier decisions, but none could return reasoning due to the nature of the ablation.

In every arm with a recovery tool, runs that made a recovery call were \kept{} more often than runs that made none (\cref{tab:recovery}), by a wide margin under \TC, \vcc{} and \TCnr{} and a narrow one under \TCng. This holds even where recovery could not return reasoning containint the explicit statement of the decision. In general recovery instead returns a more complete view of how the earlier decision implicitly played out in the relationship between the agent's earlier statements and their actions in tool calls and their output and recovering them may be part of what helped those agents keep their direction.

\subsubsection{What carried the decision without its statement}
\label{sec:carriers}

\TCnr{} could not recover the passage in reasoning in which the decision was explicitly stated, yet it kept the direction in 40 of 70 runs (and 14 of 20 with Terra). The decision left implicit traces in the behaviour that followed the statement in reasoning. In the entry where it takes shape \entry{64}, the agent's reasoning ran: \textit{``if we can get a reference to the TUI instance (via \code{ctx.ui.custom((tui, \dots)}), we can wrap \code{tui.copyTextToClipboard}''}. The visible text of the same entry acted on it: \textit{``Let me check how the extension UI context is wired to the TUI instance in interactive mode.''} Its tool calls searched Pi's interactive mode for where a factory is handed the TUI, and the investigation then went on as if the route existed. \TC's checkpoint keeps that sentence and those calls verbatim.

Most \kept{} \TCnr{} runs stated the hook early, as a conditional and from that context: \textit{``If the extension can reach the TUI instance \dots, it could monkey-patch \code{copyTextToClipboard}''} (\code{ec} \entry{339}). Of the 40 \kept{} runs, 34 stated it with no recovery call before it, most shortly after fork (27 of 40; \cref{fig:timeline}). Twenty-eight runs quoted or paraphrased the \textit{``wired to the TUI instance''} sentence after the fork, and 21 of these were \kept{} (75\%), against 19 of the other 42 (45\%). This fits \cref{tab:stages}: arms whose checkpoint kept the agent's prose verbatim came to the copy-path code with the plan more often (\TC{} 70\%, \TCnr{} 54\%, \TCng{} 48\%) than arms that kept it as a digest or a summary (\vcc{} 33\%, \pidef{} 23\%). \vcc{} keeps a recent stretch of the session verbatim and reduces everything before it to a digest of the user's requests, the files read and changed, and a few sampled lines. The agent's own account of what it was doing and why is kept only for that recent stretch, so the visible text and actions that followed an earlier decision are not in the checkpoint unless they fall within it. Here the agent could see that it had read \code{tui-renderer.js}, but not what it had concluded from reading it or where that had led it.

\begin{table}[t]
	\centering
	\small
	\footnotesize
	\setlength{\tabcolsep}{4pt}
	\begin{tabular}{@{}lccccc@{}}
		\toprule
		& \TC{} & \vcc{} & \pidef & \TCng & \TCnr \\
		\midrule
		runs with any recovery call & 74\% (26/35) & 71\% (25/35) & -- (no tool) & 46\% (16/35) & 47\% (33/70) \\
		recovery calls & 37 & 95 & -- & 37 & 57 \\
		\quad aimed at exact tool output & 78\% (29) & 68\% (65) & -- & 89\% (33) & 82\% (47) \\
		\quad aimed at design or decisions & 22\% (8) & 6\% (6) & -- & 0\% (0) & 18\% (10) \\
		\quad returned the pre-fork plan & 6 & 0 & -- & 1 & 0 \\
		\kept, if any recovery call & 77\% (20/26) & 44\% (11/25) & -- & 63\% (10/16) & 70\% (23/33) \\
		\kept, if no recovery call & 44\% (4/9) & 10\% (1/10) & -- & 53\% (10/19) & 46\% (17/37) \\
		re-read \code{tui-renderer.js} & 77\% (27/35) & 86\% (30/35) & 74\% (26/35) & 77\% (27/35) & 74\% (52/70) \\
		\kept, if re-read & 85\% (23/27) & 40\% (12/30) & 50\% (13/26) & 70\% (19/27) & 69\% (36/52) \\
		\kept, if not re-read & 13\% (1/8) & 0\% (0/5) & 0\% (0/9) & 13\% (1/8) & 22\% (4/18) \\
		\bottomrule
	\end{tabular}
	\caption{Recovery calls and re-reads after the fork (Experiments~1, 3 and~4). These are the number individual calls, not the number of runs in which a call occurred -- some runs involved several calls.}
	\label{tab:recovery}
\end{table}

\subsubsection{How the agent reflected on its own compacted state}

Four patterns recur in how agents monitored their own memory after compaction:

\paragraph{Treating the record as its own, with status attached.} As their guidance instructed, \TC{} runs typically opened by re-orienting (\textit{``Re-orientation\dots{} \code{git status}\dots{} re-confirm''}). They treated claims carried over from before compaction as claims to be checked: \textit{``From my earlier investigation (pre-compaction, now needs verification)''} (\code{q8} \entry{340}); \textit{``According to my earlier reading (compacted)\dots{} Let me re-verify by looking''} (\code{q0} \entry{339}); \textit{``I think, but I'm not 100\% sure''} (\code{r0} \entry{342}). In Experiment~1, seven \TC{} and four \vcc{} runs went looking directly for the decision that they had concluded before compaction, while no \pidef{} run showed any awareness that the hook had been decided before the fork.

\paragraph{Taking a second-hand account for its own earlier work.} Under \pidef, most agents knew they were working from a second-hand record: 9 of 35 mentioned compaction explicitly, and 22 more referred to the summary or to their earlier work as something reported to them. Twelve doubted a particular claim in it and checked it against the code or by re-running a test, almost always the summary's account of trailing spaces, which a test could settle. The summary's verdict that the copy could not be hooked was treated differently: agents took it up as settled, and in the clearest cases as their own finding. \code{a0} rejected the hook \textit{``as established''} \entry{337}; \code{c5} wrote \textit{``the extension has no access to the TUI instance. Confirmed.''} \entry{353}; \code{c3} wrote \textit{``(I confirmed earlier there's no public API to intercept \code{copySelection}\dots)''} \entry{357}; and \code{c6} wrote \textit{``Actually, I recall from the summary: `\dots{} An extension cannot intercept or modify what the built-in mouse selection copies.' So, from the public API, the extension can't intercept the built-in selection''} \entry{339}. This is a confusion of source. The model as summariser wrote a judgement, and the model as agent then interpreted it as something it had established, even when it named the summary as the place it recalled it from. Knowing that the record as a whole was second-hand did not lead the agents to track where each claim in it had come from.

\paragraph{Noticing a gap without recovering it, or believing an incomplete record to be complete.} Several agents saw that something was missing and could not get it back or simply failed to use their available recovery tool. \textit{``Let me look at the earlier evidence again. Actually, I can't --- it's in the compacted history''} (\pidef{} \code{c8} \entry{391}), where no recovery was available. In another: \textit{``that investigation has been compacted; the conclusion was probably\dots''} (\vcc{} \code{y6} \entry{561}), concluding with the agent guessing rather than attempting to recall. 
Under \TC, with a recovery tool at hand, \code{q3} named \textit{``the important question I was working on before compaction: how does the mouse-selection path work''} \entry{339}, then reconstructed its findings from the checkpoint rather than asking for them, reaching a different conclusion and yielding a \notkept{} result.

 The converse problem was also present. \vcc's \code{z3} walked recall ranges across the whole span of the decision. It got back the greps and reads, but not the conclusions drawn from them, and wrote \textit{``Now I have the complete context''} \entry{365} before building a keyboard shortcut. In \vcc's \code{x1} \entry{963}, a partial recall result led the agent to believe that something verified before the fork had \textit{``never actually been verified''}.
It also manifested as agents credited the post-compaction context with details it did not contain (\code{w7} \entry{357}, \textit{``the summary mentions a \code{select} event\dots''} -- it did not).

 \paragraph{Self-reflection is common, and does not by itself separate outcomes.} Awareness about compaction was near-universal across the board. Pi announces every compaction, so reflection is not surprising. But the guidance changed when agents reflected, and what their reflection led them to look for. Exploratory checks suggested that agents without it reflected on the compaction later after the fork. Early reflection gives an agent the chance to recover its plan before it reaches Pi's copy-path code, which is where outcomes divided (\cref{fig:timeline}). With the guidance, seven \TC{} runs went looking for their earlier conclusions and asked about the design; in six the route came back, and five took it up at once, all \kept. No \TCng{} run looked for the pre-fork decision: their lookups asked only for commands, tool output or earlier findings. So although removing the guidance did not lower the keep rate by enough for our test to detect (H4), it changed how agents dealt with their compacted state: when they turned to it, and what they went back for.
 
\section{Discussion}
\label{sec:discussion}
The preregistered results are straightforward. Without compaction, the agent kept its direction in every run, which shows what an undisturbed relationship to its task looks like in this case. The measure ranks the arms accordingly (H3): the uncompacted agent first, \TC{} closest to it, and the other compactors well below. On that reading, \TC{} best preserved the agent's relationship to its task, and recovered roughly half of what the alternatives lost. Experiments~3 and~4 showed that simple ablations don't isolate a specific feature of \TC{} that secures this advantage.

\subsection{Two carriers of a decision}
\label{sec:two-carriers}

\begin{figure}[t]
\centering
\includegraphics[width=\linewidth]{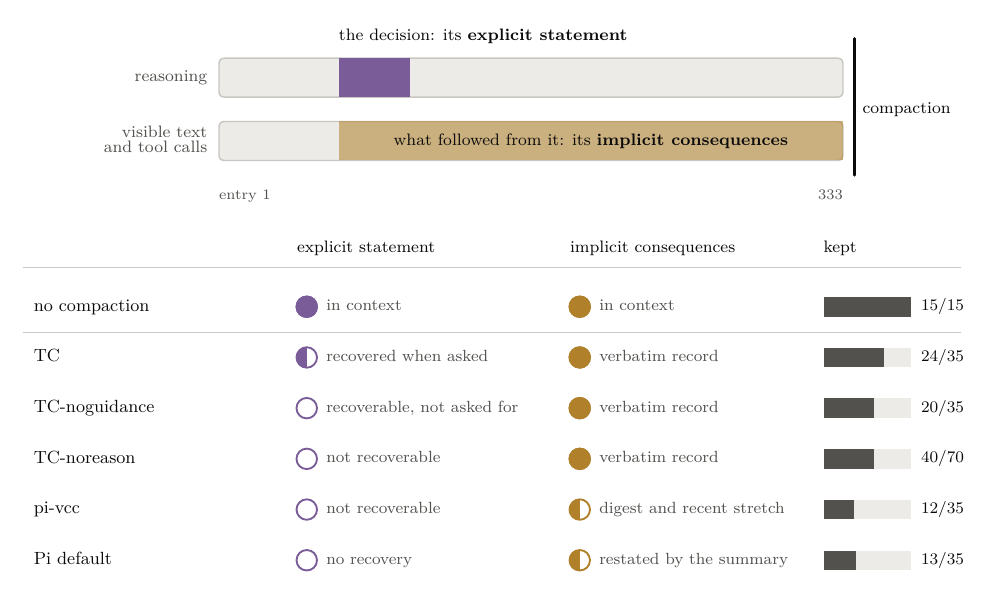}
\caption{Two carriers of the pre-compaction decision. Above: its explicit statement in the agent's reasoning, and its implicit consequences in the visible text and tool calls that followed, up to the compaction. Below: which carrier reached the agent after compaction, by arm (filled: reached it; half: on request, or in reduced form; empty: did not), with the keep rate.}
\label{fig:carriers}
\end{figure}

A decision made before compaction is carried in two ways (\cref{fig:carriers}). The first is its \emph{explicit statement}: the moment in the agent's reasoning where it decides \entry{64, 100}. The second is its \emph{implicit consequences}. A decision shapes what follows it, and all the text and actions between the decision and the compaction carry its effect. Here the same entry's visible text set out to check how the extension's UI context is wired to the TUI, and the investigation then proceeded with the understanding that route existed (\cref{sec:carriers}).

Read this way, the results fit together. The uncompacted agent has both carriers, and kept its direction every time. \TC{} keeps the implicit consequences in its verbatim record, and lets the agent recover the explicit statement on request: five runs that asked about the design got the decision back and took it up at once. \TCng{} and \TCnr{} keep the same record, but lose the explicit statement either structurally or as a second order effect on behaviour. Both landed similarly between the full \TC{} and the comparators. \vcc{} and \pidef{} lose the explicit statement too, and keep less of the implicit consequences: a digest that said nothing about where the agent was heading, or a summary that restated the consequences as conclusions, mostly negative ones. This is an interpretation, not a tested claim, but it is consistent with every result here.

In most \kept{} runs, the effect of the implicit carrier was most clearly observable at the point of re-reading Pi's copy-path code in \code{tui-renderer.js}. The code was the same in every arm, but the same lines supported opposite conclusions in different runs (\cref{sec:carriers}). Agents that came to it with the plan in mind, carried across compaction, kept their direction in 86\% of runs; those that came without it, and so had to work the plan out from the code alone, only kept it in 44\%. The arms with the verbatim record arrived with the plan more often, and among them full \TC, with its guidance and recovery of reasoning, did so most often (\cref{tab:stages,fig:timeline}). The re-read made it possible for an agent rebuild the plan from Pi's code, but whether it did so depended more on what the post-compaction context had already carried to it.

The kind of information selection that the compactor did did not decide the outcome on its own. In \pidef{} the selection was reasonable on what the summariser could see. Its rewording was the problem: every summary restated the agent's open enquiry as a negative finding, and agents then remembered that finding as their own. \vcc{} and \TC{} both select by structure in advance and defer selection by content until the need is known. But that alone did not help: \vcc{} kept the direction no more often than \pidef{} (12 and 13 of 35). What helped lay in the form of the record, and in what the agent could recover from it.

\paragraph{What this says about transparency.} \TC's transparency is not only its guidance. It has three parts: a record that shows the agent its own past words and actions as they were, rather than a retelling of them; guidance that tells the agent the record is partial, how to read it, and that the full history can still be queried; and recovery that reaches the raw session, reasoning included, and labels what it returns by source. The record is the element common to \TC{} and both ablations, all three of which kept the direction more often than the alternatives (\cref{fig:keep}). It is the most plausible carrier of the implicit consequences, and of most of \TC's advantage.

\paragraph{Why the implicit carrier matters beyond this case.} When the explicit statement was delivered, it was always effective in carrying the pre-compaction decision. This is seen in the five \TC{} runs that asked for their earlier design and got it back took it up at once, all of which kept their direction. However an agent's prior reasoning is often not available. Many models, hosted ones in particular, hide their reasoning or show only a summary of it (\cref{sec:intro}), and a harness such as Claude Code does not keep it in readable form, so for them a decision's explicit statement is lost at compaction whatever the compactor does, and carrying the implicit consequences is of particular practical benefit. And a decision need not be stated explicitly at all. A verbatim record of what followed the decision remains when the statement is not retrieved, not recoverable, or never written down, and Experiment~4 shows that it can carry the decision on its own. Experiment~5 shows the same on a different, hosted model: \TCnr{} kept the direction in 14 of 20 runs, and \pidef{} in none. Of the methods compared here, only \TC{} kept both carriers.

\subsection{Transparency as a metacognitive stance}
\label{sec:cogsci}

We describe the agent's states in the vocabulary of belief and intention as an interpretive stance on its behaviour \citep{dennett1987intentional}, not as a claim about what goes on inside the model; whether language models can introspect at all is itself disputed \citep{singh2026introspect}. In that vocabulary, compaction disrupts the agent's knowledge of its own state. The decision is an intention the agent had adopted, and after compaction the agent has to come to know that intention again. The two carriers of \cref{sec:two-carriers} are two routes to that knowledge. Recovering the explicit statement gives the agent its own earlier account of the decision. Inferring the intention from what it went on to say and do is plan recognition turned on itself \citep{mirsky2021introduction,baker2009action}: what the record keeps determines what can be inferred. Both routes depend on the agent recognising that it needs to take them, and this is where the guidance acts.

Guidance is the part of \TC's transparency that concerns the agent's knowledge of its own compacted context. Mechanical compaction with retrieval keeps part of the agent's history in its context and lets it query the rest. \TC{} also tells the agent about the state of that context: that part of its past has been removed, that what it carries are earlier claims rather than verified facts, and how to recover what it needs. This is metacognitive information, about how the agent knows what it knows rather than about the task, and it directly instructs the agent to take a \emph{metacognitive} stance towards its compacted context. The guidance is not the only source of that stance. Pi announces every compaction, and the record itself shows the agent its own words with the tool outputs and reasoning cut from between them, so the gap is visible whether or not it is explained. That structure is the record, and cannot be ablated without removing it. Without the guidance, agents still reflected on compaction, and they still kept their direction in most runs, helped by the record. But unlike in the full version of \TC, they never asked about the design, so the explicit statement of the decision, which their recovery tool could still have returned, stayed out of reach. In the terms of metamemory research, this is an effect on control: with the same recovery tools, both able to return reasoning, 8 of 37 calls asked about the design or earlier decisions with the guidance and none did without it (\cref{sec:exploratory}). The guidance does not tell the agent what to ask. It tells the agent what kind of gap it has. Recovery tools and post-compaction contexts have to work together. A tool the agent has no reason to call affords nothing.

An analogy from vision makes the idea concrete. Each eye has a blind spot, where the optic nerve leaves the retina, yet we see no hole there: the brain fills it in from the surroundings. The patch looks like the rest of the scene, so nothing gives us reason to doubt it, and a small object that falls there simply does not show up to us. A summary fills the place of the history in the same way: what it supplies is plausible and fits what surrounds it, and nothing in it marks where the detail is missing. Clinical evidence of patients with hemianopia, the loss of half the field of vision, shows how metacognitive awareness shapes behaviour around a blind spot \citep{warrington1962completion}: Shown half a circle, patients unaware of their loss see a whole one, while those who know where they cannot see actively compensate and accurately report only the half that is there. The two groups differ not in what they can see, but in whether they know the gap is there. A summary leaves the agent in the first group; \TC's guidance moves it towards the second.

The analogy also reverses a common design intuition. A summary tries to make compaction unobtrusive, so as not to disrupt the agent: it takes the place of the history without saying what was lost from the progression of the agent's behaviour, which leaves the process opaque to the agent. If the hypothesis is right, that opacity may exaggerate the harm compaction does, and the better course is, perhaps counterintuitively, to show the agent its own past as it was, and tell it plainly that its state has changed, what was removed and how to recover it. This is the sense in which \TC{} is \emph{transparent}. Claude Code takes a different course. Its guidance is explicit in part: it tells the agent that a summary has replaced the earlier conversation, that the summary may be out of date, and where the full transcript can be read. But it then asks the agent to carry on ``as if the break never happened'' \citep{claudecode2026}. That may suit the strong frontier models it is built for, which may be able to work out for themselves how far to trust a summary and when to check; in our earlier study, \TC's advantage over Pi's default summary in retaining facts and commitments eroded with a frontier model, GPT-6.1 Sol \citep{woolford2026transparent}. A weaker model may need to be told more plainly what kind of gap it has. This is speculative: here, removing \TC's guidance specifically did not significantly lower the keep rate (H4), though it did change how the agents behaved with respect to their recovery tools. How much guidance contributes, and for which models, needs further work to clarify.

\paragraph{Cognitive-science analogues.} The stance has direct analogues in the study of metamemory, monitoring what one knows and controlling recall on that basis \citep{nelson1990metamemory}, and of source monitoring \citep{johnson1993source}. Both sides of metamemory appear in the runs. \TC{} agents more often marked carried-over claims as needing verification, while \pidef{} agents more often treated a summariser's verdict as \textit{``established''}, a feeling of knowing attached to something they had never established. The \pidef{} agent that wrote \textit{``I confirmed earlier''} about a summariser's claim made a classic source-monitoring error, and \TC's provenance labels and re-orientation message are designed to address this: they separate what was said from what was verified. Some language models can draw on their earlier intentions to tell their own outputs from text inserted into their context \citep{lindsey2026emergent}; after compaction by summary, much of the text in the agent's context is not its own, and nothing in it says so. The closest analogue is choice blindness. When participants' chosen option is covertly swapped for the one they rejected, most fail to notice, and go on to justify a choice they never made \citep{johansson2005failure}. The \pidef{} agents that defended a summariser's negative verdict as their own conclusion behaved in much the same way, and nothing in the summary marked the verdict as the summariser's rather than the agent's own. A swap that is marked can be noticed and questioned.

\subsection{What this case suggests for design and evaluation}
\paragraph{Design for invariance of the task relationship, not for retention.} No compactor retained the decision, and none was designed to: all three remove reasoning. Measured by retention of that fact in the post-compaction context, the arms would tie at zero. \TC{} still did best. \citet{sun2026compactbench} reach the same point from the other side: in their repair tasks, summaries that kept the target file and most or all of the code strings needed for the fix still lost repairs that the full history completed. Retention is neither necessary nor sufficient for carrying on. We suggest that compactors be designed and evaluated against this target: after compaction the agent should relate to its task as it did before, and tests should probe that relationship as well as recall. Factual probes \citep{factory2025compression,woolford2026transparent} remain necessary, but they test what the agent knows about its task, not its awareness of where it stands in it: whether it still holds its own resolutions as settled, and its conclusions as its own. The changes users most complain of (\cref{sec:intro}), and those seen here, are of that kind.

\paragraph{What should carry across, in goal-directed terms.} The user reports (\cref{sec:intro}) point to features of the pre-compaction state that matter for continuing the task: the \emph{direction}, the \emph{status} of each item, \emph{attribution}, the agent's \emph{own voice}, and \emph{resolutions together with the doubts they resolve}. Only the first was tested here, and the exploratory analysis helps illustrate the others. The last is the clearest lesson of the no-compaction runs. The doubt is easy to re-derive from the code, and the resolution is not, so a compactor that removes both should at least say that a question was settled, and where to find how. A record that keeps the structure and form of the agent's own words and actions keeps a decision's consequences even when its statement is lost. Recovery should reach wherever decisions are formed, which for reasoning models is often in reasoning, and keep what was observed apart from what it infers, so that the agent can weigh a negative bottom line against the facts beneath it (\code{ei}; \cref{sec:exploratory}).

\paragraph{What a repeatable task like this is good for.} The test case trades generality for fine-toothed analysis over repeated instances. It is one sharp probe, cheap enough to repeat many times, and it shows both how often and \emph{why} a run loses or recovers its direction, which broad benchmarks of agentic coding \citep[e.g.][]{jimenez2024swebench} and full-length twin sessions rarely expose. Ideally it would be one of a suite of cases, each built around a different feature of the task relationship, e.g. a lost negative, a change of status, a question of attribution, and a commitment that cannot be re-derived so easily from disk. Candidates can be found by scanning sessions at their first compaction for conclusions that are later contradicted. There is work to be done on how to make this search efficient. So far we have found only this one case, along with a handful of similar but less sharp examples.

\section{Conclusion}
\label{sec:conclusion}

Users of compacting agents report that something other than facts goes missing: direction, status, attribution, tone. We tested one part of that relationship, its direction, on a natural case. A real coding-agent session was forked at the moment its first compaction intervened between the agent choosing an approach and building it, and run forward many times, changing only the compactor. In pre-registered tests, under \TC{} the agent carried on in its pre-compaction direction about twice as often as under \vcc{} or Pi's default summary (24/35 against 12/35 and 13/35). With compaction switched off it carried on in 15 of 15 runs, so the measure ranks the uncompacted agent first and \TC{} closest to it. 

In every arm, the agent lost its direction mostly in the same way: after the fork it came to believe that Pi's copy could not be hooked, reopening a doubt it had already resolved, sometimes on the word of a misleading summary. Whether it kept its direction depended mostly on whether it came back to Pi's code with the plan already in mind. Runs that did so kept their direction in 86\% of cases, against 44\% for runs that read the same code without it. Runs arrived with the plan most often under \TC{} and least often under Pi's default summary, and in a few \TC{} runs the plan came back directly, when the agent asked for its earlier design and got it.

Removing \TC's guidance did not significantly lower the \kept{} rate (H4), although it changed what the agent looked for: without it, no run asked for its earlier design. Without access to reasoning, \TC{} still kept the direction significantly more often than the other methods (H5), so its advantage does not rest on reasoning alone. On a different model, GPT-6.1 Terra, \TC{} without access to reasoning kept the direction in 14 of 20 runs, and Pi's default in none.

\paragraph{}

We interpret the evidence as pointing towards two ways that the decision gets carried across the compaction boundary. Its explicit statement, written in reasoning, reaches the agent only if the agent has structural access to the reasoning text and the knowledge required to reach for it.
Its implicit consequences are carried in the agent's own words and actions that followed the statement. 
It appears that \TC's transparency allows it to carry both more effectively than the compared methods in this case. 
In metacognitive terms, \TC{} supports the agent's knowledge of its own state along both routes: its guidance tells the agent what kind of gap its memory has, which shapes what it tries to recall, and its record keeps the agent's earlier words as its own, which shapes what it can infer.
How much each part contributes is a question for a future dedicated ablation study across multiple cases.

\section*{Author contribution and use of AI}
\addcontentsline{toc}{section}{Disclosure of AI involvement}

FW designed the study, its hypotheses, its interpretations, and \code{transparent-compact-pi}. He used AI tools (Claude Code, Pi with OpenAI and Qwen models) to assist in organising and carrying out the experiment: running and monitoring the sessions, computing the statistics, and parsing the session text for the exploratory qualitative analyses. The blind scores were produced by separate model instances, which saw only redacted packets and the rules. FW wrote the manuscript, with AI (Claude) assistance in phrasing in some sections, tracking of arguments and data, and \LaTeX{} and plot formatting. KS secured funding and resources, contributed to the interpretations, and provided feedback on the manuscript.

\section*{Replication package}
\addcontentsline{toc}{section}{Reproducibility statement}

The replication package, at \repro, holds the frozen fork snapshot, the build and check scripts, the compactor builds, the scoring rules, the blind scorers scripts, and the redacted scoring packages for the runs presented here.

\bibliographystyle{plainnat}
\bibliography{refs}

\appendix
\clearpage
\section{Scoring rules and outcomes by rule}
\label{app:rules}

Defined before any study run was recorded. Rules were defined by analysing precedents in runs from pilot sessions and the source session.

\paragraph{Kept: the committed extension hooks Pi's own mouse-selection copy, so an ordinary selection copies clean.}
\begin{description}[leftmargin=0.9cm,labelwidth=0.7cm,font=\normalfont\bfseries]
  \item[A1] Any method of reaching the TUI counts (e.g.\ a zero-line \code{setWidget}, a momentary \code{ui.custom} overlay, capture at session start).
  \item[A2] Any point on the copy path counts: wrapping the injected \code{copySelection}, \code{getActiveSelectionText}, \code{copyTextToClipboard} or \code{copySelectionToClipboard}, on the instance or the prototype, by assignment or \code{defineProperty}.
  \item[A3] Hooking the gesture instead of the method counts (e.g.\ wrapping terminal input and overwriting Pi's copy on mouse release).
  \item[A4] How the text is cleaned doesn't matter. Falling back to the stock text when it can't map is fine.
  \item[A5] Partial coverage counts if the main case is covered (e.g.\ only selections inside the last reply, or an added command alongside the hook).
  \item[A6] The report doesn't override the code.
\end{description}
\paragraph{Not kept.}
\begin{description}[leftmargin=0.9cm,labelwidth=0.7cm,font=\normalfont\bfseries]
  \item[B1] A separate key or command after the stock copy, even with full TUI access.
  \item[B2] The reason for not hooking doesn't matter (a wrong belief, a deliberate design choice, ``not a normal extension'').
  \item[B3] Working on stored text with no TUI access (e.g.\ a \code{/copylines} picker over the message text).
  \item[B4] Only committed work counts.
  \item[C1] A hook that installs only when the user runs a command first.
  \item[C2] A hook that is committed but plainly can't fire, judged by reading the code.
  \item[C3] Hooking a copy path other than mouse selection (e.g.\ only Ctrl+X).
  \item[C4] A hook on a copy mechanism other than Pi's own (e.g.\ tmux's \code{copy-pipe}).
\end{description}

\Cref{tab:rules} breaks the outcomes down by deciding rule. \emph{Kept} runs almost all wrapped a method on the copy path (A2). \emph{Not-kept} runs most often worked on stored text with no TUI access (B3), or added a separate shortcut after the stock copy (B1).

\begin{table}[h]
\centering
\small
\begin{tabular}{@{}lccccccc@{}}
\toprule
arm & A2 & A3 & B1 & B3 & B4 & C3 & C4 \\
\midrule
\TC{} 0.11.1 & 23 & 1 & 3 & 7 & 1 & 0 & 0 \\
\vcc{} 0.11.1 & 11 & 1 & 7 & 11 & 3 & 1 & 1 \\
\pidef & 12 & 1 & 4 & 16 & 2 & 0 & 0 \\
\midrule
NC & 15 & 0 & 0 & 0 & 0 & 0 & 0 \\
\midrule
\TCng & 20 & 0 & 4 & 9 & 2 & 0 & 0 \\
\TCnr & 40 & 0 & 8 & 20 & 1 & 1 & 0 \\
\midrule
\TCnr{} (Terra) & 14 & 0 & 1 & 5 & 0 & 0 & 0 \\
\pidef{} (Terra) & 0 & 0 & 7 & 13 & 0 & 0 & 0 \\
\bottomrule
\end{tabular}
\caption{Deciding rules, by arm: Experiment~1, the no-compaction reference of Experiment~2, the ablations of Experiments~3 and~4, and Experiment~5 on GPT-6.1 Terra. A = \kept; B, C = \notkept.}
\label{tab:rules}
\end{table}
\FloatBarrier

Of the 56 runs of Experiment~1 that were \notkept, most committed a design that works on stored text with no access to the TUI (B3, 34 runs) or a separate command after the stock copy (B1, 14). Six committed nothing within the turn (B4), and two hooked a copy path or mechanism other than Pi's mouse copy (C3, C4). Nine runs reached the 90-minute cap and were scored, as registered, on what they had committed. In hindsight this rule slightly favoured \TC. Two capped \TC{} runs (\code{r6}, \code{r7}) had already committed a \kept{} hook, while five capped runs (\TC{} \code{r0}, \vcc{} \code{x6} and \code{y1}, \pidef{} \code{b2} and \code{b6}) had committed nothing but were judged from their reasoning to be on track to commit the \kept{} solution. The rule did not affect any result (Appendix \ref{app:capping}).

\clearpage
\section{Alternative scoring of runs stopped at the cap}
\label{app:capping}

Nine runs of Experiment~1 reached the 90-minute cap and were scored, as pre-registered, on what they had committed (Appendix~\ref{app:rules}). By chance, this turned out to give a small advantage to TC, which had 2/3 sessions \kept{} under this rule, while the other arms had all capped sessions end without any commitments. The cap was chosen to prevent time wastage on runaway sessions, but does not necessarily reflect the effectiveness of the compaction method. This advantage did not affect the significance of the results: \Cref{tab:capping} gives every test under the three defensible alternate scorings of those runs.

\begin{table}[h]
\centering
\small
\footnotesize
\setlength{\tabcolsep}{4pt}\begin{tabular}{@{}lcccccc@{}}
\toprule
capped runs & \kept: \TC{}/\vcc{}/\pidef & H1 & H2 & H3a & H3b & H3c \\
\midrule
as registered & 24 / 12 / 13 & 0.0028 & 0.0052 & 0.0010 & $2.2 \times 10^{-6}$ & $6.8 \times 10^{-6}$ \\
(a) \code{r6}, \code{r7} \notkept{} & 22 / 12 / 13 & 0.011 & 0.021 & 0.0003 & $2.2 \times 10^{-6}$ & $6.8 \times 10^{-6}$ \\
(b) all seven on-track runs \kept{} & 25 / 14 / 15 & 0.0045 & 0.0087 & 0.0016 & $1.1 \times 10^{-5}$ & $2.1 \times 10^{-5}$ \\
\bottomrule
\end{tabular}
\caption{Every test under two alternative scorings of the runs stopped at the 90-minute cap: (a) the two capped \TC{} runs that had committed a \kept{} hook scored \notkept; (b) all seven capped runs that post-hoc analysis deemed on track to the \kept{} solution (\code{r0}, \code{r6}, \code{r7}, \code{x6}, \code{y1}, \code{b2}, \code{b6}) scored \kept. Keep counts for Experiment~1's arms; NC is 15/15 throughout. Every null is still rejected.}
\label{tab:capping}
\end{table}

\clearpage
\section{Blind scoring by other models}
\label{app:scorers}

After the scoring method was registered, we considered that using the same model three times was a potential limitation. As a check, used four other models to score H1--H5 in the same way: the same packets, rules and prompt, each set freshly shuffled, with access to nothing but the packet directory. Three were hosted models: GPT-6.1 Sol, GPT-6.1 Terra and GPT-6 Luna. The fourth was Qwen3.8-27B, the study's own developing model, run locally with the study's settings. \Cref{tab:scorers} gives each model's agreement with the registered score. The three Opus scorers had agreed with each other on every packet.

\begin{table}[h]
\centering
\footnotesize
\begin{tabular}{@{}lcccccc@{}}
\toprule
scorer & H1 & H2 & H3 & H4 & H5 & all \\
\midrule
GPT-6.1 Sol & 70/70 & 70/70 & 30/30 & 50/50 & 85/85 & 305/305 \\
GPT-6.1 Terra & 70/70 & 69/70 & 30/30 & 50/50 & 84/85 & 303/305 \\
Qwen3.8-27B & 70/70 & 68/70 & 29/30 & 49/50 & 85/85 & 301/305 \\
GPT-6 Luna & 55/70 & 51/70 & 30/30 & 38/50 & 63/85 & 237/305 \\
\bottomrule
\end{tabular}
\caption{Agreement of four other models with the registered score (the majority of three Opus scorers) on \kept{} or \notkept, by packet set. Each set includes the runs mixed in from Experiment~1.}
\label{tab:scorers}
\end{table}

Agreement follows the generally understood ability ranking of the scorer model: Sol agreed on every packet, Terra and Qwen on all but two and four, while Luna disagreed with more. The disagreements point the same way. Every one was a \kept{} run scored \notkept. None of Terra's or Qwen's would change a registered result, and in five of the six the scorer's reason was a factual claim about Pi's runtime that Pi's source contradicts. The sixth, \code{p8}, polls the selection and overwrites the clipboard, and sits on the boundary of the rules. Luna's disagreements came from failing to find the code. Its search skipped hidden directories, where most runs keep their extension, and it then scored the remaining runs without opening them. Departures from the Opus scores therefore appear to reflect the limits of the model doing the scoring rather than uncertainty in the scores. Where the models agreed on \kept{} or \notkept, they mostly gave the same rule. The most common difference was between B1 and B3, a boundary on which the Opus scorers also split.

\clearpage
\section{Effort and cost}
\label{app:cost}

\begin{figure}[h]
\centering
\includegraphics[width=0.9\linewidth]{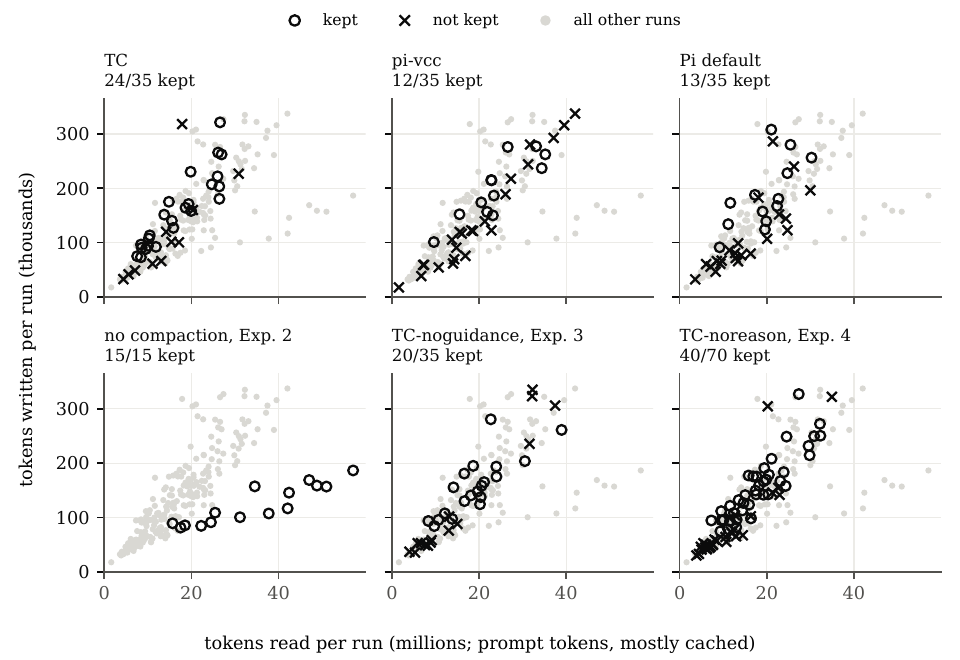}
\caption{Token cost of every run, by arm: tokens read (prompt tokens summed over the developing agent's model calls, most of them served from the prompt cache) against tokens written (output tokens). Circles are \kept{} runs, crosses \notkept; grey points are the runs of all other arms.}
\label{fig:tokens}
\end{figure}

Transparency was not expensive here. Neither \TC{} nor \vcc{} makes a model call to compact, whereas \pidef{} spent 3.9M input and 0.56M output tokens on its summaries across 35 runs, and \TC's lookups generated at most an estimated 2.3\% of the developing agent's output tokens. The main hidden cost on local hardware was the occasional loss of the prompt cache when the lookup subagent ran (about 8 minutes across the arm). \Cref{tab:effort} shows effort and cost per run after the fork, and \cref{fig:tokens} the token cost of every run against its outcome. Without compaction, the agent read about twice as many tokens per run as under any compactor, since every call carries the whole history.

It is noteworthy that although \TC's checkpoints are larger than the other compactors, it ultimately led to less prompt and output tokens being used across the turn, although the difference between \TC{} and \pidef{} is minor.

\begin{table}[h]
\centering
\small
\footnotesize\setlength{\tabcolsep}{4pt}
\begin{tabular}{@{}lccccc@{}}
\toprule
arm & minutes & model calls & compactions & prompt tokens & output tokens \\
\midrule
\TC{} 0.11.1 & 35 (8--90) & 166 (61--304) & 2 (1--4) & 15.4M (4.4--30.8M) & 127k (33--322k) \\
\vcc{} 0.11.1 & 46 (4--90) & 222 (32--477) & 2 (1--5) & 21.3M (1.6--42.0M) & 150k (18--338k) \\
\pidef & 40 (8--89) & 171 (60--334) & 2 (1--4) & 17.3M (3.6--30.3M) & 124k (32--308k) \\
\midrule
NC & 59 (42--117) & 134 (71--192) & 0 & 34.5M (15.7--57.1M) & 109k (81--187k) \\
\midrule
\TCng & 38 (10--90) & 181 (56--393) & 2 (1--5) & 16.6M (4.0--38.9M) & 130k (35--335k) \\
\TCnr & 32 (8--90) & 150 (50--332) & 2 (1--5) & 14.0M (3.9--34.9M) & 110k (30--327k) \\
\bottomrule
\end{tabular}
\caption{Effort and cost per run after the fork, median (range): Experiment~1, the no-compaction reference of Experiment~2, and the ablations of Experiments~3 and~4. Prompt tokens are summed over the developing agent's model calls. Peak context, the largest single prompt, was about 159k in every compaction arm, and 297k (median) under NC. NC ran on the slower 512k set-up (\cref{sec:h3-design}), so its minutes are not directly comparable. Runs capped at 90 minutes: \TC{} 3, \vcc{} 4, \pidef{} 2, \TCng{} 2, \TCnr{} 3.}
\label{tab:effort}
\end{table}

\end{document}